\documentclass[lettersize,journal]{IEEEtran}
\usepackage{amsmath,amsfonts}
\usepackage{algorithmic}
\usepackage{algorithm}
\usepackage{array}
\usepackage[caption=false,font=normalsize,labelfont=sf,textfont=sf]{subfig}
\usepackage{textcomp}
\usepackage{stfloats}
\usepackage{url}
\usepackage{verbatim}
\usepackage{graphicx}
\usepackage{cite}
\usepackage{booktabs}
\usepackage{multirow}

\usepackage{amsmath}
\usepackage{amsthm}
\usepackage{amsfonts}
\usepackage{amssymb}
\usepackage{bm}
\usepackage{mathrsfs}
\usepackage{mathtools}
\usepackage{dsfont}

\usepackage{thmtools}
\declaretheorem{definition}

\declaretheorem{theorem}
\declaretheorem{lemma}
\declaretheorem{proposition}

\declaretheorem[style=remark]{remark}
\declaretheorem{example}

\def\cA{{\mathcal{A}}}

\def\cC{{\mathcal{C}}}
\def\cD{{\mathcal{D}}}

\def\cF{{\mathcal{F}}}
\def\cG{{\mathcal{G}}}

\def\cI{{\mathcal{I}}}

\def\cP{{\mathcal{P}}}

\def\cU{{\mathcal{U}}}

\def\cX{{\mathcal{X}}}
\def\cY{{\mathcal{Y}}}

\def\scA{{\mathscr{A}}}

\def\scU{{\mathscr{U}}}

\def\bbN{{\mathbb{N}}}

\def\bbR{{\mathbb{R}}}

\DeclareMathOperator*{\argmin}{arg\,min}

\newcommand{\E}{\mathbb{E}}
\newcommand{\op}[1]{\operatorname{#1}}

\newcommand{\RPE}{\text{RPE}}

\newcommand{\TF}{\operatorname{TF}}

\newcommand{\ID}{\text{ID}}

\newcommand{\Attn}{\text{Attn}}
\newcommand{\POLA}{\text{POLA}}
\newcommand{\supp}{\operatorname{supp}}

\newcommand{\SRC}{\operatorname{SRC}}

\usepackage{xcolor}

\newcommand{\teal}[1]{\textcolor{teal}{#1}}
\newcommand{\orange}[1]{\textcolor{orange}{#1}}

\def\APPDXHAUSDORFF{A}
\def\APPDXNFLLG{B}
\def\APPDXPEIMP{C}
\def\APPDXEXPERIMENTS{E}

\begin{document}

\title{On the Limitations and Capabilities of Position Embeddings for Length Generalization
\thanks{© 2025 IEEE. Personal use of this material is permitted. Permission 
from IEEE must be obtained for all other uses, in any current or future 
media, including reprinting/republishing this material for advertising or 
promotional purposes, creating new collective works, for resale or 
redistribution to servers or lists, or reuse of any copyrighted 
component of this work in other works.}
}

\author{Yang Chen, Yitao Liang, and Zhouchen Lin
\thanks{Y. Chen is with the State Key Laboratory of General Artificial Intelligence, School of Intelligence Science and Technology, Peking University, Beijing, China (e-mail: yangchen@pku.edu.cn).}
\thanks{Y. Liang is with the Institute for Artificial Intelligence, Peking University, Beijing, China, and also with the Beijing Institute for General Artificial Intelligence, Beijing, China (e-mail: yitaol@pku.edu.cn).}
\thanks{Z. Lin is with the State Key Laboratory of General Artificial Intelligence, School of Intelligence Science and Technology, Peking University, Beijing, China, and also with the Institute for Artificial Intelligence, Peking University, Beijing, China, and with Pazhou Laboratory (Huangpu), Guangzhou, Guangdong, China (e-mail: zlin@pku.edu.cn).}
\thanks{Corresponding to: Z. Lin and Y. Liang.}
}



\maketitle

\begin{abstract}
    In Transformers, Position Embeddings (PEs) significantly influence Length Generalization (LG) performance, yet their fundamental role remains unclear. In this work, we investigate the limitations and capabilities of PEs in achieving LG. We theoretically analyze PEs in Position-Only Linear Attentions (POLAs), introducing Linear Representation Complexity (LRC) to characterize when PEs enable LG. Our analysis shows that PEs do not expand computational capabilities but structure learned computations across positions. Extending to practical Transformers, we propose Sequential Representation Complexity (SRC) and conjecture that LG is possible if and only if SRC remains invariant across scales. We support this hypothesis with empirical evidence in various reasoning tasks. To enhance LG, we introduce Scale Hint, allowing flexible instance scaling, and a Learning-Based Position Embedding framework that automatically learns positional relations. Our work provides theoretical insights and practical strategies for improving LG in Transformers.
\end{abstract}


\section{Introduction}\label{sec:introduction}

Length Generalization (LG) refers to the ability of a model to extrapolate from small-scale instances to larger ones in reasoning \cite{anil2022exploring, zhou2023algorithms, zhou2024transformers, huang2025formal}. In many tasks, the sample space grows exponentially with the problem scale, making exhaustive training infeasible. Thus, it is important to learn from limited training samples at small scales while generalizing to larger ones. Furthermore, learning to solve complex tasks from simple ones is a significant ability of human learning. LG is an essential aspect when building a model of human-level reasoning capability \cite{lake2017building, bahdanau2019systematic, lake2023human}.

In general, LG is inherently difficult because the training data do not provide information on how to compute results for unseen large-scale instances. No single algorithm can guarantee length generalization across all tasks \cite{chen2025low} (see Appendix~\APPDXNFLLG 
for a further illustration). As a result, incorporating prior knowledge about the target concept is crucial for designing effective models. Current works have adopted various techniques to encode such prior knowledge. In Transformers, Position Embeddings (PEs) are found to play a significant role in LG performance \cite{kazemnejad2024impact, zhou2024transformers}. However, few works have theoretically investigated why and how PEs enable LG \cite{huang2025formal}. Moreover, certain tasks fail to generalize with existing PEs \cite{zhou2023algorithms, jelassi2023length, hahn2024sensitive}. It is unclear whether this is due to suboptimal PE strategies or fundamental PE limitations. A fundamental question arises naturally:
\begin{center}
    \emph{What are the limitations and capabilities of PEs for LG?}
\end{center}

A PE encodes positional relations between elements in a sequence. Intuitively, these relations define how the model interprets the positional structure of a sequence, specifying how positions interact and influence computations, enabling the model to distinguish between different positions and determine positional dependencies. The relations are determined by a Positional Relation Function (PRF), denoted as $\phi(i,j)$, which maps a query position $i$ and a key position $j$ to a value that represents their relationship. For example, in Relative Position Embedding (RPE) \cite{shaw2018self}, the function is $\phi(i,j)=i-j$; in Absolute Position Embedding (APE) \cite{vaswani2017attention, devlin2019bert}, the PRF can be seen as $\phi(i,j)= i * K +j$ for some constant $K$ such that $\phi(i_1,j_1)\neq\phi(i_2,j_2)$ for any $0\leq j_1\leq i_1\leq N-1$, $0\leq j_2\leq i_2\leq N-1$, $(i_1,j_1)\neq(i_2,j_2)$, where $N$ is the maximum length considered. Some PEs have very different implementations but they share the same PRF and thus capture the same position relations. For instance, while learnable RPE \cite{shaw2018self} and RoPE \cite{su2024roformer} implement positional relations with learnable vectors and rotary matrix respectively, they have the same PRF $\phi(i,j)=i-j$. See Appendix~\APPDXPEIMP 
for a more detailed illustration. This work focuses on the role of positional relations in LG and will mainly discuss the impact of PRFs.

It is challenging to analyze the impact of PEs on the LG in practical Transformers. It is necessary to take the learning process into account when it comes to LG. Representation capabilities are insufficient when considering LG. For example, Transformers equipped with APE are proved to be Turing-complete \cite{perez2021attention}, which are expressive enough for all computable problems; in LG, however, Transformers with APE typically have poor LG performance \cite{shaw2018self, jelassi2023length, kazemnejad2024impact}. Furthermore, since practical Transformers are typically overparameterized and the training data only provide imperfect information, the LG performance might depend on the inductive bias of the training algorithm \cite{abbe2023generalization, hahn2024sensitive}. However, analyzing the learning process can be extremely difficult for practical Transformers due to their nonlinearity and high complexity.

\subsection{Our Works}

\textbf{Limitations and Capabilities of PEs in Position-Only Linear Attentions.}
To isolate the role of PEs in LG, we first analyze a simplified Transformer variant called Position-Only Linear Attention (POLA, Definition~\ref{def:pola}). In these models, the linear attention scores depend only on positional relationships. In POLAs, different PEs can be seen as different linear reparameterizations of attention matrices. By studying POLAs, we can theoretically examine how PEs contribute to generalization before extending our insights to full Transformers.

We focus on the LG of the tasks within the expressive power of the POLA models. To characterize the tasks whose LG can be achieved or not by PEs, we define Linear Representation Complexity (LRC, Definition~\ref{def:lrc}), which quantifies the number of independent computational patterns (or ``operators'') needed to represent a task within a POLA model. Specifically, LRC of a POLA model on a domain is defined as the size of the minimal set of disjoint $\{0,1\}$-valued matrices that can linearly combine the attention sub-matrix restricted to the domain. 

We first show that PE cannot help LG for the tasks (apart from a negligible subset) whose LRC strictly increases when shifting from the training domain to the testing domain. Intuitively, the negative result means that PEs cannot help to learn new ``operators'' beyond the training data. On the other hand, we prove that LG can be achieved with a proper PE for the POLA whose LRC in the testing domain is invariant to that in the training domain. More specifically, when we choose the PE that makes the positions with the same operators share the same learnable parameters, we can achieve LG by training the POLA model with gradient descent.

\textbf{Limitations and Capabilities of PEs in Practical Transformers.} 
Our analysis of POLAs reveals a key insight: PEs may not introduce new computational capabilities but rather help the model determine which operations to apply at different positions. This suggests that in practical Transformers, PEs may not expand the range of functions the model can learn, but they can help ensure that learned computations are applied consistently as sequence length increases.

We propose Sequential Representation Complexity (SRC) to characterize the task complexity, how many distinct computational components (``operators'') are required for solving a reasoning task. We conjecture that LG can be achieved by adapting the PE only if the SRC remains the same when the task scales up. Furthermore, when the SRC is invariant, we conjecture that choosing a PE that correctly identifies the positions for the operators can promote LG.

We provide empirical evidence supporting the conjecture (Section~\ref{sec:practical-tf}). For the scenario where SRC increases, we fail to achieve LG by solely adapting PEs. In various reasoning tasks where SRC does not increase, we achieve LG by solely adapting PEs, making PRFs identify the operators.

In one word, we establish a fundamental boundary: \emph{PEs cannot introduce new operators beyond training data, but they can consistently identify and apply the same operators across scales to enable LG}.

\textbf{Scale Hint Technique.} 
Using a single PRF to identify positions can be overly restrictive, as it requires the instances to have a fixed structure across \emph{all} scales. For example, in the Addition task, to achieve LG with RPE, we might need to align the addends and the sum to the target length. Fixing structures may also lead to computational inefficiency. When we train with addition within 5 digits, we might need to insert padding 0s to align the addends to the target length, namely 20 digits, which leads to a heavy extra computation cost.

To alleviate the problem, we introduce a technique called Scale Hint (SH) in Section~\ref{subsec:sh}. 
If we know the scales of the instances, we can incorporate the instance scale as an input to the PRF. Then we only need a fixed structure within each scale, rather than enforcing a single structure across all scales. This allows a more flexible data format and also potentially reduces the computational cost. For instance, when dealing with an $n$-digit Addition instance, we only need to align each addend to $n$ digits rather than the target length with the PRF 
\begin{equation*}
    \phi(i,j,n) = K \lfloor (i - j) / n\rfloor + \min\left((i - j) \mod n, K - 1 \right),
\end{equation*}
where $i$ is the query position, $j$ is the key position, $n$ is the scale hint, and $K$ is some constant. 

\textbf{Learning-Based Position Embeddings.}
In practice, it would be unrealistic to manually redesign the PRF task by task. It is desirable to use a single model across different tasks. To address this problem, we propose learning-based PE (Section~\ref{subsec:lbpe}), respectively, where the PRF $\phi$ is learned automatically. More concretely, we replace the handcrafted PRF $\phi(i,j)$ ($\phi(i,j,n)$ if scale hint is employed) with a learnable one $\phi_{\theta}(i,j)$ ($\phi_{\theta}(i,j,n)$, respectively). Empirically, we observe that the learning-based PEs achieve length generalization across a variety of tasks, eliminating the need for task-specific designs. This approach shows the potential to use one learnable PE to handle diverse tasks that would otherwise require different handcrafted designs.

\begin{figure*}[th]
    \centering
    \subfloat[IPE]{%
        \includegraphics[width=0.32\linewidth]{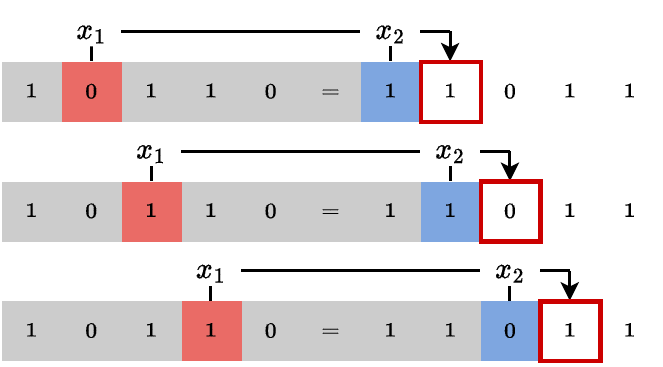}
    }%
    \hfill
    \subfloat[APE]{%
        \includegraphics[width=0.32\linewidth]{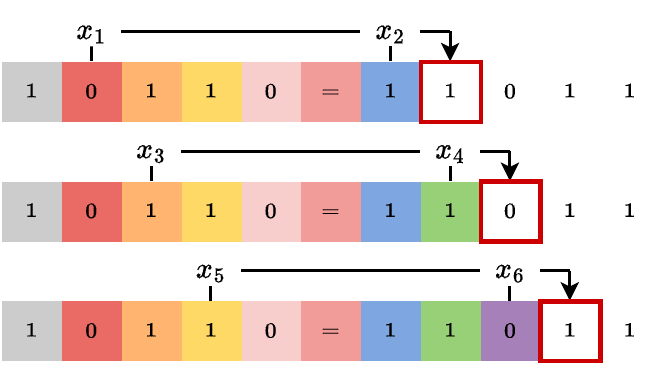}
    }%
    \subfloat[RPE]{%
        \includegraphics[width=0.32\linewidth]{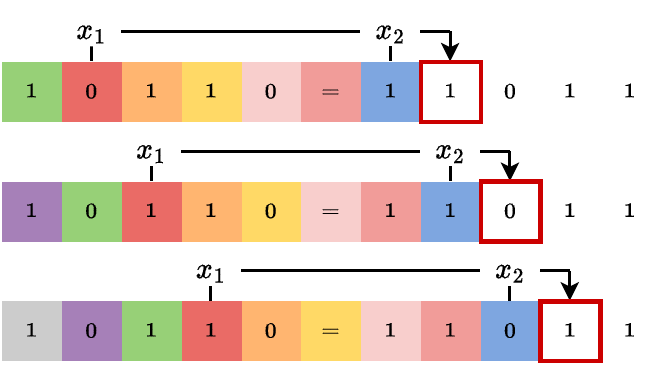}
    }%
    \caption{Different PEs correspond to different methods for computing outputs in the Parity (with CoT) task. IPE (see Section~\ref{sec:practical-tf}) and RPE align the positions across steps and scales to compute the next token from the corresponding token in the input ($x_1$) and the current token ($x_2$). IPE encodes all other positions into a single value, whereas RPE redundantly encodes them with distinct values. In contrast, APE lacks positional alignment, requiring a distinct operator at each step. When input scales exceed those seen during training, APE necessitates novel operators not learned from data. Under the notation introduced in Section~\ref{subsec:anc}, both IPE and RPE characterize a circuit of non-increasing SRC that computes the Parity task, while APE does not. As we show, PEs alone cannot introduce novel operators or handle circuits with increasing SRC. Consequently, IPE and RPE succeed in achieving LG, while APE fails to generalize.}
    \label{fig:illustration-ipe-ape-rpe}
\end{figure*}
\section{Related Work}\label{sec:related_work}

\textbf{Length Generalization in Reasoning Tasks.}
The literature on length generalization can be broadly categorized into two strands. The first focuses on \emph{processing extremely long sequences}, often referred to as \emph{long-context modeling} in the context of LLMs \cite{han2024lm, hu2024longrecipe, li2024functional, fang2025wrong, yuan2025native}. This line of research primarily addresses the challenges of capturing long-range dependencies and mitigating the substantial memory and computational demands associated with long inputs.

The second line of work, to which the present study contributes, investigates \emph{generalization from shorter training sequences to longer sequences at inference time} \cite{jelassi2023length, huang2025formal}. The central question is how models can generalize by learning from length-limited data that provide only incomplete information in training. Addressing this question necessitates the introduction of suitable inductive biases \cite{abbe2023generalization, chen2025low}, which may be incorporated through data formatting \cite{zhou2023algorithms, xiao2025generalizing}, architectural modifications \cite{shaw2018self, teney2024neural}, or training strategies \cite{li2018algorithmic, lyu2024dichotomy}. This work specifically examines the role of PEs, a structural component of Transformers, as a means of facilitating LG.

\textbf{Role of Position Embeddings in Transformers.}
PEs encode positional information that is otherwise absent in the standard Transformer architecture. They are critical for enabling Transformers to model sequences in a position-sensitive manner and have been shown to significantly enhance the model's expressive capacity. In particular, it has been demonstrated that Transformers are Turing-complete when equipped with APE, but not without them \cite{perez2021attention}.

Recent empirical studies have highlighted the importance of PEs in length generalization \cite{kazemnejad2024impact, zhou2024transformers}. Replacing APEs with PEs that encode relative position relations has been shown to improve performance on tasks requiring extrapolation to longer sequences \cite{shaw2018self, su2024roformer}. Moreover, more sophisticated PEs that encode structured or hierarchical positional information can lead to further gains \cite{he2018two, cho2024position, mcleish2024transformers}. Despite these promising empirical findings, a comprehensive theoretical understanding of the role that PEs play in enabling length generalization remains limited. This work takes a step towards such understanding by providing a systematic theoretical investigation into how modifications to PEs influence generalization behavior across sequence lengths.

\section{Preliminary}\label{sec:preliminary}

In this work, we consider the following definition for LG.

\begin{definition}[$(N_0,N)$-Length Generalization]\label{def:lg}
    Suppose that in an instance $x=c_1,\dots,c_n\in\Sigma$ of length $n$, each element $c_i$ is sampled i.i.d. from some distribution $\cP$ on $\Sigma$, where $\Sigma$ is the element domain and the support set of the distribution $\cP$ is $\Sigma$, i.e., $\supp(\cP)=\Sigma$. A learning algorithm $\scA$ achieves $(N_0,N)$-Length Generalization ($(N_0,N)$-LG) for the concept $f^*$ if there exists a distribution $\cP_{N_0}$ on $[N_0]$ such that the model $f$ learned by $\scA$ on a sufficiently large $\cX_{N_0}=\{x_k\}$, which is generated by $(n, x)\sim \cP_{N_0}(n)\cP(x\mid n)$ satisfies $f(x)=f^*(x)$ for all $x\in\Sigma^n$, $n=1,\dots,N$.
\end{definition}

Definition~\ref{def:lg} captures two core challenges in length generalization: (1) the exponential growth of the input space with increasing length, and (2) the absence of information in shorter instances about components unique to longer inputs. In some cases, it is more convenient to fix the length of $x$ and include an additional indicator $n$ representing the ``effective length'', the portion of $x$ that contributes to the output. The training data takes the form $\cX_{N_0} = {(n_k, x_k)}$, sampled from $(n, x) \sim \cP_{N_0}(n) \cP(x \mid n)$, with $x$ always of length $N$. This alternative preserves the two essential difficulties of LG.

\textbf{Notations.} We use $\cU_N$ to denote the set of all upper triangular matrices in $\bbR^{N\times N}$. We use $[N]$ to denote the set $\{1, \dots, N\}$. The cardinality of a set $A$ is denoted by $|A|$. $\Sigma^{[N]}$ denotes the set of all strings over $\Sigma$ of length at most $N$, i.e., $\Sigma^{[N]} = \bigcup_{k \in [N]} \Sigma^k$. Let $\Sigma^*$ denote the Kleene closure of the alphabet $\Sigma$, i.e., $\Sigma^* = \bigcup_{k=1}^{\infty} \Sigma^k$. We use $\mathds{1}(\cdot)$ to denote the indicator function. We use $H^d(\cdot)$ to denote the $d$-dimensional Hausdorff measure and $\dim_H(\cdot)$ to denote the Hausdorff dimension in Euclidean space (see Appendix~\APPDXHAUSDORFF 
for a brief review). We use $\Delta^A$ to denote the probability simplex over the set $A$, i.e., $\Delta^A=\left\{(x_a)_{a\in A}\in\bbR^A\mid x_a \ge 0, \sum_{a\in A}x_a=1\right\}$.
\section{Warmup: A Study in POLAs}\label{sec:pola}

To understand how PEs influence LG, we begin by studying a simplified variant of the Transformer architecture, termed Position-Only Linear Attention (POLA). The POLA model isolates positional relationships, allowing us to theoretically analyze the intrinsic capabilities and limitations of PEs before extending insights to practical Transformers.

\begin{definition}[Position-Only Linear Attention]\label{def:pola}
    A Position-Only Linear Attention (POLA) model computes an output from an input sequence $x\in\bbR^{N}$ and a positional indicator $n\,(n\leq N)$ as follows:
    \begin{equation*}
        f_{\POLA}(x,n;A) = x^\intercal A e_n = \left\langle x e_n^\intercal, A\right\rangle,
    \end{equation*}
    where $A\in\cU_N$ is the learnable parameter and $e_n\in\bbR^N$ denotes the vector with a $1$ in the $n$-th coordinate and $0$'s elsewhere.
\end{definition}

We now illustrate how the POLA model is simplified from a standard linear attention.
Given the input sequence $x$, a standard linear attention computes the output at the query position $n$ as follows:
\begin{equation*}
    f_{\Attn}(x,n;\Theta) = \sum_{i\leq n} \left[\left(W_Q x_n\right)^\intercal \left(W_K x_i\right) + B_{n,i}\right]  W_V x_i,
\end{equation*}
where the learnable parameters $\Theta=\left\{W_Q, W_K, W_V, B\right\}$ include query, key, value transformations, and position bias $B$. To isolate positional relationships, we remove the token-related attention terms, specifically $\left(W_Q x_n\right)^\intercal \left(W_K x_i\right)$. Furthermore, since our focus is length generalization, we treat the value transformation $W_V$ as non-pivotal and thus fix it (specifically to the identity matrix $I$ without loss of generality). The intuition is straightforward: If a model cannot learn an appropriate value transformation from training data, it would fail even at basic in-distribution generalization, making length generalization considerations irrelevant. These simplifications make our analysis clearer and specifically highlight the role of positional relations in length generalization.

In the POLA model, each position embedding can be seen as a linear reparameterization of the attention matrix $A$, i.e., $A=\sum_{s=1}^S U_s^p q_s$, where $(U_s)_{i j}$ indicates whether the positional relation between query position $i$ and key position $j$ equals $s$ according to a PRF $\phi(i,j)$, i.e.,
\begin{equation*}
    \left(U_s\right)_{ij}=\begin{cases}
        1, & \phi(i,j)=s,\\
        0, & \text{otherwise}.
    \end{cases}
\end{equation*}
For example, RPE can be viewed as this linear reparameterization by choosing $U_s = D_s$, $0\leq s\leq N-1$, where $D_s$ denotes the $s$-th upper diagonal indicator matrix defined as:
\begin{equation*}
    \left(D_s\right)_{i j} = \begin{cases}
        1, & i - j = s,\\
        0, & \text{otherwise}.
    \end{cases}
\end{equation*}

To characterize when POLA models with PEs can achieve LG, we define Linear Representation Complexity (LRC).

\begin{definition}[Linear Representation Complexity]\label{def:lrc}
    Suppose that $A\in\bbR^{d_1\times \dots \times d_R}$ is tensor. Then $A$ can be represented as a linear combination of a set of tensors $\cU=\{U_1,\dots,U_K\}$, i.e., $A=\sum_{k=1}^K a_k U_k$ for some $a_1,\dots,a_K$, where $(U_k)_{i_1,\dots, i_R}\in\{0,1\}$ and $(U_{k_1})_{i_1,\dots, i_R}(U_{k_2})_{i_1,\dots, i_R}=0$ for $k_1\neq k_2$. Denote the set of all sets satisfying the above condition by $\scU$. The Linear Representation Complexity (LRC) of the tensor $A$, represented by $\op{LRC}(A)$, is the size of the minimal set in $\scU$:
    \begin{equation*}
        \op{LRC}(A) = \min_{\cU\in\scU} \left|\cU\right|.
    \end{equation*}

    For a given POLA model $f^*$, its LRC on domain $\cX$ up to the input length $N$ is the minimum LRC among all parameters $A$ satisfying $f_{\POLA}(x,n;A) = f^*(x,n)$ for all $(x,n) \in \cX\times [N]$. Formally,
    \begin{equation*}
        \op{LRC}(f^*;\cX,N) := \min_{A\in\cA_{\cX,N}(f^*)} \op{LRC}(A),
    \end{equation*}
    where $\cA_{\cX,N}(f^*)$ is defined as:
    \begin{equation*}
         \left\{A\mid f_{\POLA}(x,n;A) = f^*(x,n) \text{ for all } (x,n)\in\cX\times [N]\right\}.
    \end{equation*}
    Furthermore, we define
    \begin{equation*}
        \op{LRC}(A;\cX,N)= \op{LRC}\left(f_{\POLA}(x,n;A);\cX,N\right).
    \end{equation*}
\end{definition}

Intuitively, LRC measures the minimal number of positional relations required to distinguish different computational roles in a task, reflecting the number of different operators needed to solve it. A higher LRC implies that more distinct operators are intrinsically necessary to solve the task. Theorem~\ref{thm:pola-limitation} formalizes a fundamental limitation of PEs for LG in terms of LRC.

\begin{theorem}\label{thm:pola-limitation}
    Define $\cF_M:=\{A\in\cU_N\mid \|A\|_{\infty}\leq M\}$ and $\cF_{M,B}:=\{A\in \cF_M\mid A_{[N_0],[N_0]}=B\}$ for all $B\in\cU_{N_0}$. 
    For any $B_0\in\cU_{N_0}$ and fixed learning algorithm, let $\cF_{M,B_0}^{N_0,N}\subseteq\cF_{M,B_0}$ be the subset consisting of all elements $A$ such that there exists a PE achieving $(N_0, N)$-LG for $f_{\POLA}(x,n;A)$ with the algorithm. Let $\tilde{\cF}_{M,B_0}^{N_0,N}\subseteq\cF_{M,B_0}$ be the subset of increasing LRC, i.e.,
    \begin{equation*}
        \tilde{\cF}_{M,B_0}^{N_0,N}:=\left\{A\in\cF_{M,B_0}\mid \op{LRC}\left(A;\cX, N_0\right)<\op{LRC}\left(A;\cX, N\right)\right\}.
    \end{equation*}
    Then for all $M > 0$, we have
    \begin{equation*}
        \dim_H\left(\tilde{\cF}_{M,B_0}^{N_0,N}\setminus\cF_{M,B_0}^{N_0,N}\right) = 
        \dim_H\left(\tilde{\cF}_{M,B_0}^{N_0,N}\right):=d_N,
    \end{equation*}
    and 
    \begin{equation*}
        H^{d_N}\left(\tilde{\cF}_{M,B_0}^{N_0,N}\setminus\cF_{M,B_0}^{N_0,N}\right) = 
        H^{d_N}\left(\tilde{\cF}_{M,B_0}^{N_0,N}\right).
    \end{equation*}
\end{theorem}

\begin{remark}
    Theorem~\ref{thm:pola-limitation} does not simply state that a fixed model with a fixed learning algorithm cannot distinguish a target function from others due to limited training data. It makes a stronger claim: the PE can be chosen with access to not only the training data but also \emph{any} external information, including perfect prior knowledge of the target function. Even under this idealized setting, for almost all tasks with increasing LRC, LG cannot be achieved solely by adapting the PE. Therefore, Theorem~\ref{thm:pola-limitation} reveals a fundamental limitation of PEs, one that is \emph{not} a consequence of the no-free-lunch theorem for LG.
\end{remark}

Theorem~\ref{thm:pola-limitation} indicates that adapting PEs alone cannot achieve LG for almost all tasks with increasing LRC. Intuitively, this is because PEs cannot introduce new operators for larger-scale instances. This naturally raises a complementary question: \emph{Can we achieve LG for all tasks with non-increasing LRC by selecting appropriate PEs?} Theorem~\ref{thm:pola-capability} confirms this positively:

\begin{theorem}\label{thm:pola-capability}
    For any $f^*(x,n)=f_{\POLA}(x,n;A^*)$ such that $\op{LRC}(f^*,\cX_{N_0})=\op{LRC}(f^*,\cX_N)$, then there exists a PE such that the POLA model with this PE initialized at $0$ and trained by gradient descent achieves $(N_0,N)$-LG.
\end{theorem}

Theorem~\ref{thm:pola-capability} shows that, given an appropriate PE, POLA models can achieve LG via gradient descent whenever LRC remains invariant. This suggests that the essential role of PEs is to properly utilize existing operators learned during training when scaling to larger instances.

\section{PEs for LG in Practical Transformers}\label{sec:practical-tf}

The insights from POLAs suggest that PEs can only help to identify the usage of the learned operators, promoting LG only for the scenarios where LRC does not increase. In this section, we study whether these capabilities and limitations of PEs hold in practical Transformers.

For better clarity, we introduce several concepts and notations in the next subsection.

\subsection{Additional Concepts and Notations}\label{subsec:anc}

To clearly formalize how sequence-to-sequence mappings $f: \Sigma^* \mapsto \Sigma^*$ are computed autoregressively, we introduce a formalism to represent sequential computations via circuits.

\begin{definition}[Circuit Representation of Sequential Computation]
    Consider a sequence-to-sequence mapping $f:\Sigma^*\mapsto\Sigma^*$. For an input sequence $x=(x_1,\dots,x_n)\in\Sigma^n$ mapped to an output sequence $f(x)=(y_1,\dots,y_{m(n)})\in\Sigma^{m(n)}$ for some function $m:\bbN\mapsto\bbN$, we define the \emph{circuit representation of sequential computation} as follows: 

    Let $z=(z_1,\dots,z_{n+m(n)})$ represent the concatenation of input and output sequences, i.e.,
    \begin{equation*}
        z_i = \begin{cases}
            x_i, & 1\leq i\leq n,\\
            y_{i-n}, & n+1\leq i\leq n+m(n).
        \end{cases}
    \end{equation*}

    A circuit representation $\cC_N=\{C_n\}_{n\in [N]}$ of this sequential computation up to input length $N$ is a collection of sets of tuples, where
    \begin{equation*}
        C_n = \left\{(i,g^{(i)},I_i)\mid 1\leq i\leq n+m(n)\right\},
    \end{equation*}
    where each triple $(i,g^{(i)},I_i)$ denotes that the $i$-th element $z_i$ is computed by applying an operator $g^{(i)}$ to the set of parent elements indexed by 
    \begin{equation*}
        I_i = (i_1,\dots, i_{r_i})\subseteq [i-1].
    \end{equation*}

    Formally, the computation is defined as:
    \begin{equation*}
        z_i = g^{(i)}(z_{i_1},\dots,z_{i_{r_i}}),\quad\text{ for } n+1\leq i\leq n + m(n),
    \end{equation*}
    with $g^{(i)}\in G(\cC_N)$ where $G(\cC_N)=\{g_1,\dots,g_R\}$ is the set of unit operators used in the representation, and each operator $g_r:\Sigma^{d_r}\mapsto\Sigma$ has arity $d_r\leq D$. For the input tokens ($1 \leq i \leq n$), we define $g^{(i)}=g_0$, an operator representing input gates, and set $I_i=\emptyset$.

    For simplicity of notation, we may omit the tuples representing input gates $(i, g_0, \emptyset)$ in $C_n$ for all $1\leq i\leq n$ when this omission does not cause ambiguity. We may also drop the subscripts $N$ and $n\in[N]$ when the explicit indication of the maximum input length is unnecessary.
\end{definition}

This definition formalizes the notion of sequential computation using a circuit analogy, providing clear notation to analyze how computations unfold step-by-step. The next two examples illustrate
the circuit representation of sequential computation.

\begin{example}[Parity (with CoT)]
    Consider the Parity (with CoT) task $f(x_1,\dots,x_n)=(y_1,\dots,y_n)$ where $\Sigma=\{0,1\}$, $y_1=x_1$ and $y_i = x_i \oplus y_{i-1}$ for all $2\leq i\leq n$. A circuit representation $\cC=\{C_n\}$ of this task can be given as:
    \begin{equation*}
        \begin{gathered}
            C_n = \left\{(1,g_0,\emptyset),\dots,(n,g_0,\emptyset),(n+1, g_{\text{ID}},(1))\right., \\ 
            \left.(n+2, g_{\oplus},(2, n+1)),\dots, (2 n, g_{\oplus}, (n, 2n-1))\right\},
        \end{gathered}
    \end{equation*}
    with the unit operator set $G(\cC_N)=\{g_{\ID},g_{\oplus}\}$ where $g_{\ID}:\Sigma\mapsto\Sigma$ is the identity operator, i.e., $g_{ID}(u)=u$, and $g_{\oplus}:\Sigma^2\mapsto\Sigma$ is the XOR operator, i.e., $g_{\oplus}(u, v)=u\oplus v$.
\end{example}

\begin{example}[Multiplication (1 * N)]
    Consider the Multiplication (1 * N) task $f(x_1,\dots,x_n)=(y_1,\dots,y_n)$ where $\Sigma=[9]$, $x_1$, $x_2 ... x_n$ are the multipliers, and $y_1 ... y_{n}$ is the product (the digits are reversed), satisfying:
    \begin{equation*}
        x_1 * x_n \dots x_2 = y_n \dots y_1.
    \end{equation*}
    A circuit representation $\cC=\{C_n\}$ of this task is given as:
    \begin{equation*}
        \begin{gathered}
            C_n = \left\{(1,g_0,\emptyset),\dots,(n,g_0,\emptyset),\right.\\
            \left. (n+1, g_1,(1, 2)), (n+2, g_2,(1, 2, 3, n+1)),\dots, \right.\\
            (2 n - 1, g_2, (1, n-1, n, 2 n-2)) \left.(2 n, g_3, (1, n, 2n-1))\right\},
        \end{gathered}
    \end{equation*}
    with the unit operator set $G(\cC_N)=\{g_1,g_2,g_3\}$ where
    \begin{equation*}
        \begin{gathered}
            g_1 (u_1, u_2) = u_1 * u_2 \bmod{10}, \\
            g_2 (u_1, u_2, u_3, u_4) = \left[(u_1 * u_3 \bmod{10}) + \left\lfloor u_1 * u_2 / 10 \right\rfloor\right.\\
            \left. + \mathds{1}\left( u_4 < (u_1 * u_2 \bmod{10})\right)\right] \bmod{10},\\
            g_3 (u_1, u_2, u_3) =  \left\lfloor u_1 * u_2 / 10 \right\rfloor + \mathds{1}\left( u_3 < (u_1 * u_2 \bmod{10})\right).
        \end{gathered}
    \end{equation*}
\end{example}

One sequence-to-sequence mapping $f$ can have multiple different circuit representations, even with the same operator set $G$. Different circuit representations correspond to different ways in which the mapping can be computed sequentially. When we write the function in one of its circuit representations, we specify a certain sequential computation. See the following example for an illustration.

\begin{example}\label{ex:crsc-nonuniqueness}
    Consider the mapping $f(x_1, \dots,x_n)=((x_1+1)\bmod{10}, (x_1+2)\bmod{10},\dots, (x_1 + n)\bmod{10})$, and the operator set $G={g_1, g_2}$ where $g_1(u) = (u+1)\bmod{10}$ and $g_2(u) = (u+2)\bmod{10}$. The following two circuits $\cC=\{C_n\}$ and $\cC'=\{C_n'\}$ sharing the same unit operator set represent two different ways that both compute the same function $f$:
    \begin{equation*}
        \begin{aligned}
            C_n =& \left\{(n+1, g_1, 1), (n+2, g_1, n+1), \cdots,\right.\\
            &\left.(2n-1, g_1, 2n-2),(2n, g_2, 2n-2)\right\},\\
            C'_n =& \left\{(n+1, g_1, 1), (n+2, g_2, 1),\right. \\
            &\left.(n+3, g_2, n+1), \dots,(2n, g_2, 2n-2)\right\}.
        \end{aligned}
    \end{equation*}
\end{example}

With the notion of circuit representation formally established, we can now quantify the complexity of these computations and study how PEs influence the Transformer's capability to generalize across different lengths. To this end, we introduce the concept of \emph{Sequential Representation Complexity (SRC)}.

\begin{definition}[Sequential Representation Complexity]
    Suppose that $f:\Sigma^*\mapsto\Sigma^*$ is a sequence-to-sequence mapping. We define the Sequential Representation Complexity (SRC) of $f$ up to input length $n$ as the minimal cardinality of the unit operator set among all possible circuit representations up to length $N$. Formally,
    \begin{equation*}
        \SRC(f,N) := \min_{\mathcal{C}_N \in \mathfrak{C}_N(f)} |G(\mathcal{C}_N)|,
    \end{equation*}
    where the minimization is taken over the set of all circuit representations, i.e.,
    \begin{equation*}
        \mathfrak{C}_N(f) := \{\mathcal{C}_N \mid \mathcal{C}_N \text{ is a circuit representation computing } f\},
    \end{equation*}
    and $G(\cC_N)$ denotes the set of unit operators used in $\cC_N$.
\end{definition}

SRC measures how many distinct unit operators are required to sequentially compute a sequence-to-sequence mapping up to a certain input scale. Intuitively, when the SRC at the target scale exceeds the number of operators learned in the training domain, positional information alone is insufficient to achieve length generalization. We will show this in the Section~\ref{sec:subsec:limitations}.

To formally describe the conditions under which positional information can identify the operators needed in the sequential computation, we introduce the following definition.

\begin{definition}[PRF Characterization for Circuit Representation]\label{def:prf-char}
    Consider a circuit representation $\cC=\{C_n\}$ that sequentially computes a mapping $f:\Sigma^*\mapsto\Sigma^*$. Let $\phi:\bbN\times\bbN\mapsto\bbN$ be a Position Representation Function (PRF). We say that $\phi$ \emph{characterizes} the circuit representation family $\mathcal{C}$ if it satisfies the following two conditions:
    \begin{enumerate}
        \item (Consistency) For any two tuples $(i,g^{(i)},I_i)\in C_m$ and $(i',g^{(i')}, I_{i'})\in C_n$, if
        \begin{equation*}
            \phi(i-1,j)=\phi(i'-1,j'),
        \end{equation*}
        then it must hold that
        \begin{equation*}
            g^{(i)}=g^{(i')} \quad\text{and}\quad I_i^{-1}(j)=I^{-1}_{i'}(j'),
        \end{equation*}
        where
        \begin{equation*}
            I^{-1}_{\alpha}(\beta) = \begin{cases}
                k, & \text{ if } \beta \text{ is the $k$-th element in } I_\alpha,\\
                +\infty, & \text{otherwise}.
            \end{cases}
        \end{equation*}
        
        \item (Distinctness) For any distinct pairs $(i,j)\neq (i',j')$, if there exists $C_m\in\cC, (i,g^{(i)},I_i)\in C_m, j\in I_i$ such that one of the following conditions holds:
        \begin{enumerate}
            \item there exists some $C_n\in\cC, (i',g^{(i')},I_{i'})\in C_n, j'\in I_{i'}$ such that 
            \begin{equation*}
                g^{(i)}\neq g^{(i')} \text{ or } I_{i}^{-1}(j)\neq I_{i'}^{-1}(j');
            \end{equation*}
            \item for all $C_n\in\cC$, if $(i',g^{(i')},I_{i'})\in C_n$, then $j'\not\in I_{i'}$;
        \end{enumerate}
        then it must be that
        \begin{equation*}
            \phi(i-1,j)\neq\phi(i'-1,j').
        \end{equation*}
    \end{enumerate}
\end{definition}

Intuitively, when a PRF characterizes a circuit representation, we can identify the operators and their inputs via the PRF values: identical PRF values imply identical computational roles (consistency), and different computational roles must have distinguishable PRF values (distinctness). Our experiments in Section~\ref{sec:subsec:capabilites} show that a PE can enable LG for a reasoning task when its PRF characterizes some circuit representation of the task.

With the above concepts and notations, we now investigate both limitations and capabilities of PEs for LG.

\subsection{Limitations of PEs}\label{sec:subsec:limitations}

In this section, we establish a fundamental limitation of PEs for LG when SRC grows from training to testing. Intuitively, PEs alone cannot help the model acquire operators beyond those learnable from the given training data. Consequently, if solving a larger-scale instance necessarily requires additional operators not present in the training scale, achieving LG becomes nearly impossible if only adapting PEs.

More concretely, consider the scenario where a Transformer model equipped with adaptive PEs is to learn a Boolean function up to scale $N$, having only been trained on smaller-scale instances within scale $N_0$. We prove that even with optimal choice of PEs---selected with precise knowledge of the target function---the model cannot achieve LG for nearly all functions whose SRC increases from the training scale to the testing scale. Formally, we have the following theorem:

\begin{theorem}\label{thm:limitation}
    Let $\cF_N$ be the set of all Boolean functions $f:\{0,1\}^{[N]}\mapsto\{0,1\}^K$. Consider an arbitrary but fixed Transformer architecture $\TF$ sufficiently expressive to represent any function in $\cF_N$ given an appropriate PE. Suppose that $N_0:\bbN\mapsto\bbN$ is a function such that $N_0(n)< n$ for all $n\in\bbN$ (with slight abuse of notation, we also write $N_0$ to denote $N_0(N)$ in the theorem). For any learning algorithm, let $\cF_{N_0, N}\subseteq\cF_N$ be the subset consisting of all functions in $\cF_N$ whose $(N_0, N)$-LG can be achieved by the Transformer $\TF$ with a proper PE and the algorithm. Define $\tilde{\cF}_{N_0,N}:=\left\{f\in\cF_N\mid \SRC_{N_0}\left(f\right)<\SRC_N\left(f\right)\right\}$, representing functions whose SRC strictly increases scaling from $N_0$ to $N$.
    Then we have
    \begin{equation*}
        \lim_{N\to\infty}\frac{\left|\tilde{\cF}_{N_0,N}\setminus\cF_{N_0,N}\right|}{\left|\tilde{\cF}_{N_0,N}\right|} = 1.
    \end{equation*}
\end{theorem}

Theorem~\ref{thm:limitation} implies that even under ideal conditions---optimal PE adaptation and full knowledge of the target function---PE-based approaches fail to extrapolate in length for almost all functions whose SRC grows from the smaller training scale to the larger testing scale. Thus, this reveals a fundamental limitation of employing PEs alone: without additional inductive biases or methods for acquiring new operators, achieving general LG is theoretically impossible in almost all cases of increasing SRC.

\subsection{Capabilities of PEs}\label{sec:subsec:capabilites}

While Section~\ref{sec:subsec:limitations} has demonstrated fundamental limitations of PEs in enabling LG, in this section, we investigate the scenarios where PEs indeed facilitate effective length generalization. Specifically, we show that when the target function can be solved entirely by operators already acquired in the training domain, carefully designed PEs can achieve LG. The key is that a proper PE aligns the computational ``roles'' of elements consistently across instances of different lengths. This alignment enables models to correctly identify and sequentially apply learned operators, resulting in accurate computations at larger scales.

To empirically validate this insight, we conduct experiments comparing three types of PEs across various reasoning tasks: Ideal PE (IPE), APE, and RPE. IPE is the PE whose PRF faithfully characterizes the circuit representation of the minimum unit operator set. APE and RPE are the most common PEs as the baselines in the experiments.

\begin{figure}
    \centering
    \subfloat[Copy]{%
        \includegraphics[width=0.48\linewidth]{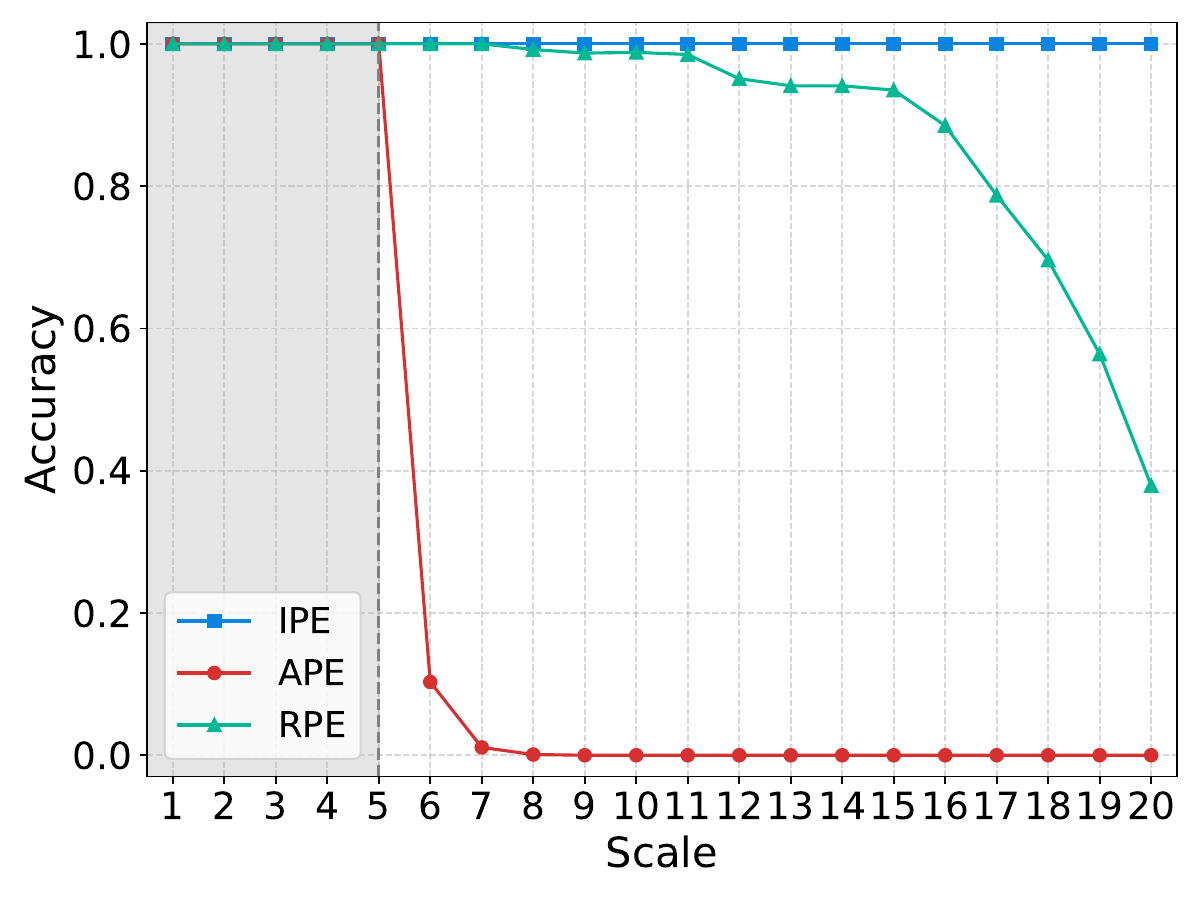}
    }
    \hfill
    \subfloat[Shift]{%
        \includegraphics[width=0.48\linewidth]{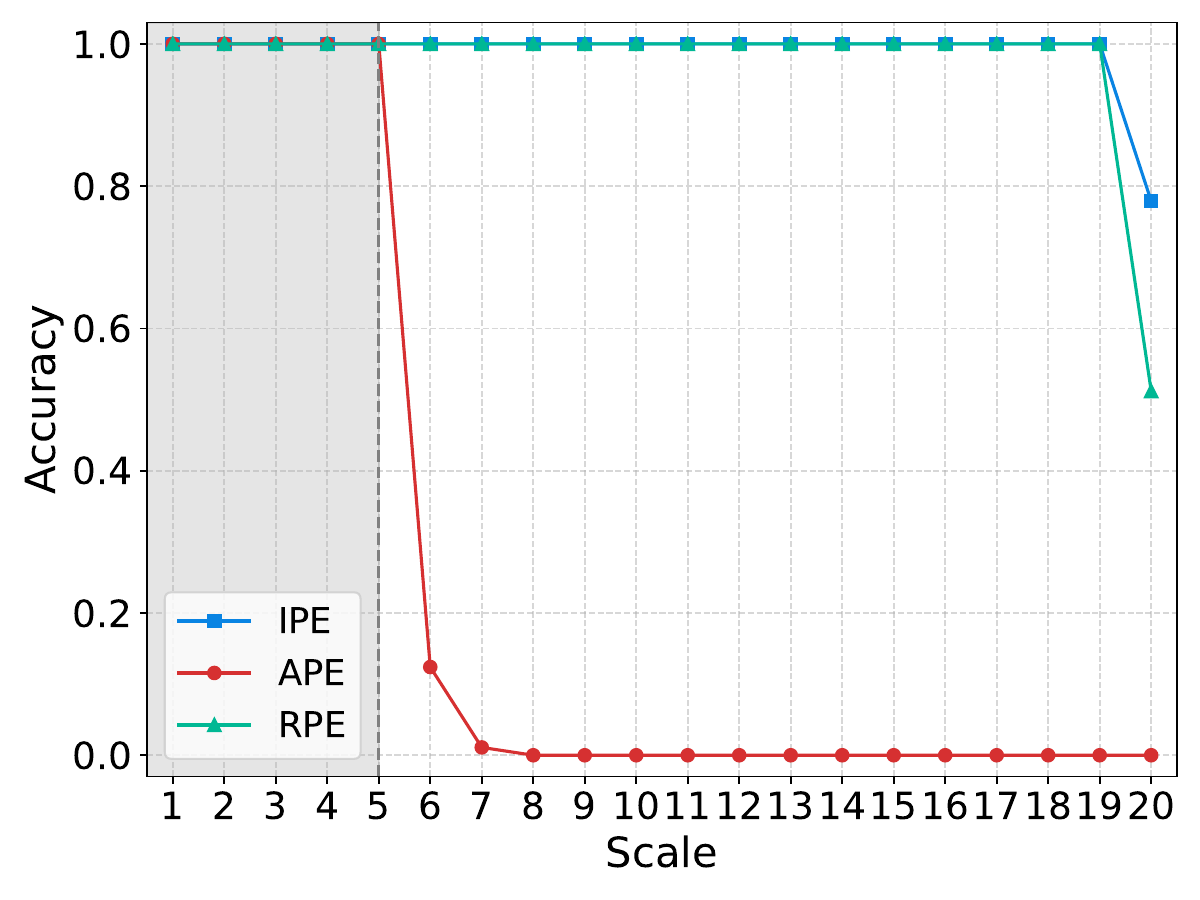}
    }%
    \\
    \subfloat[Parity (with CoT)]{%
        \includegraphics[width=0.48\linewidth]{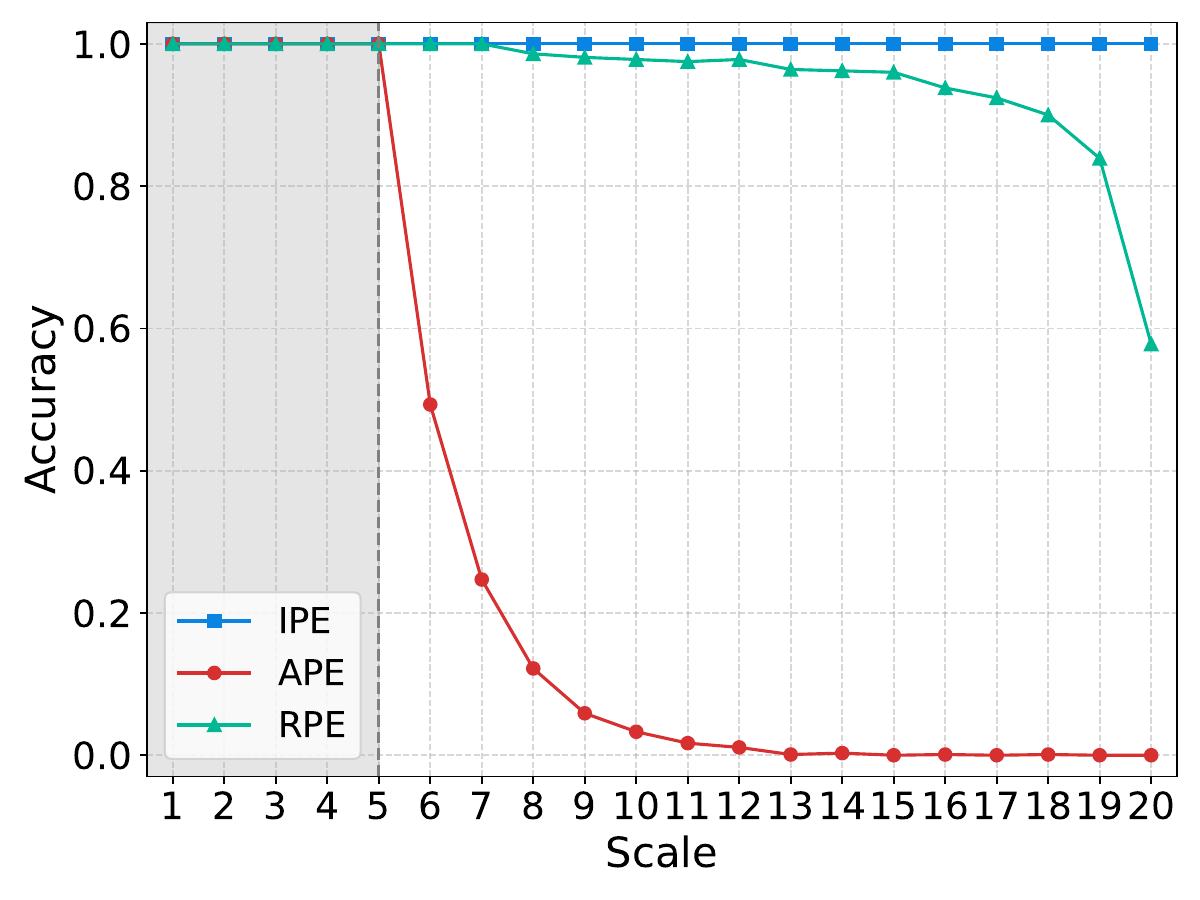}
    }%
    \hfill
    \subfloat[Addition]{%
        \includegraphics[width=0.48\linewidth]{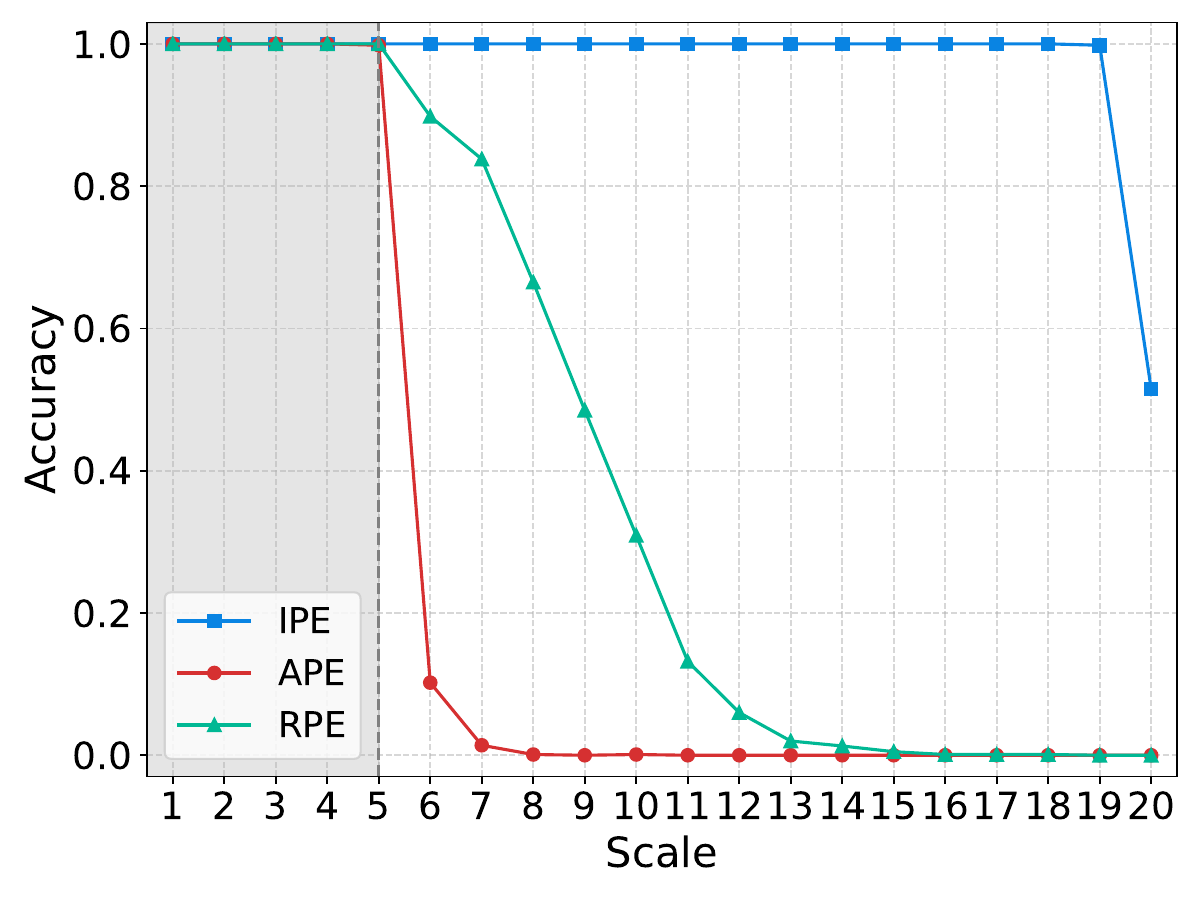}
    }%
    \\
    \subfloat[Multiplication (1 * N)]{%
        \includegraphics[width=0.48\linewidth]{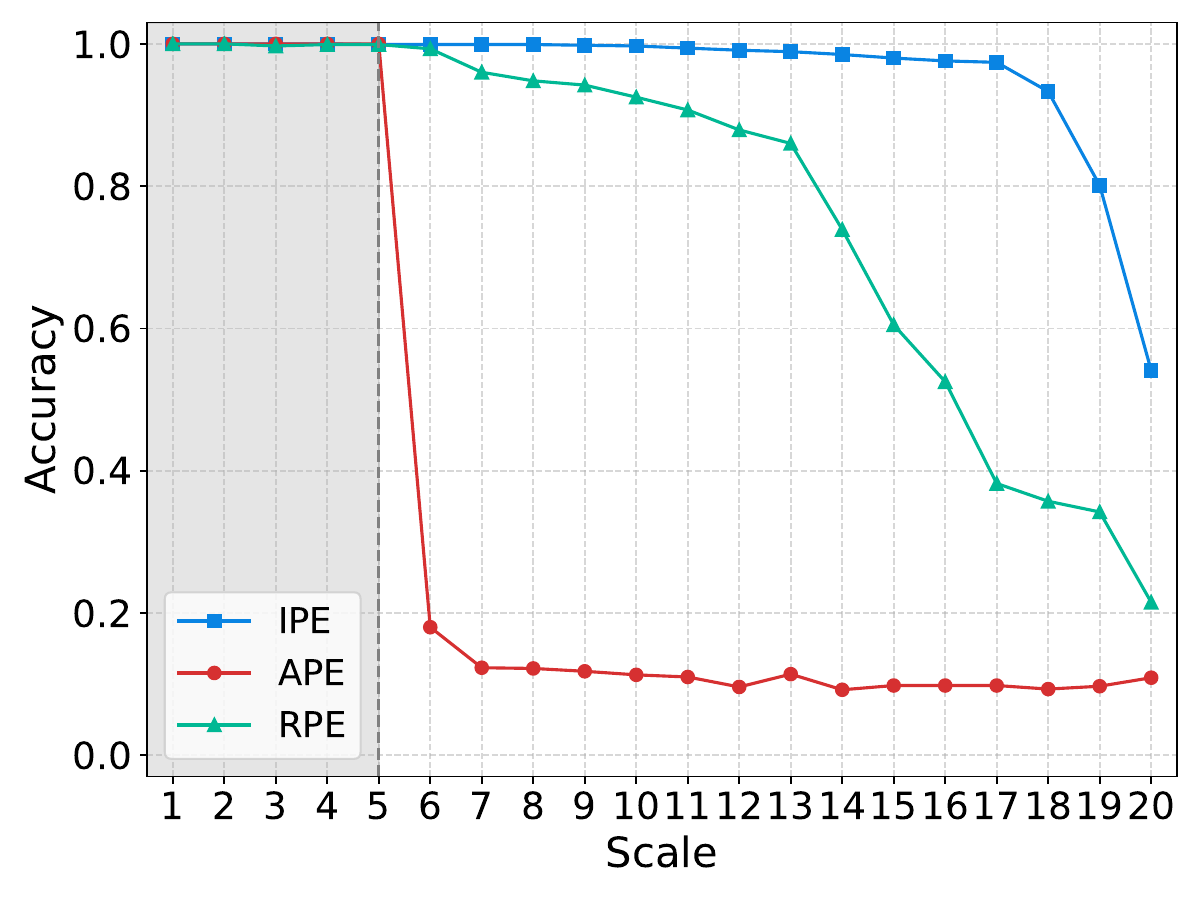}
    }%
    \hfill
    \subfloat[Division (N / 1)]{%
        \includegraphics[width=0.48\linewidth]{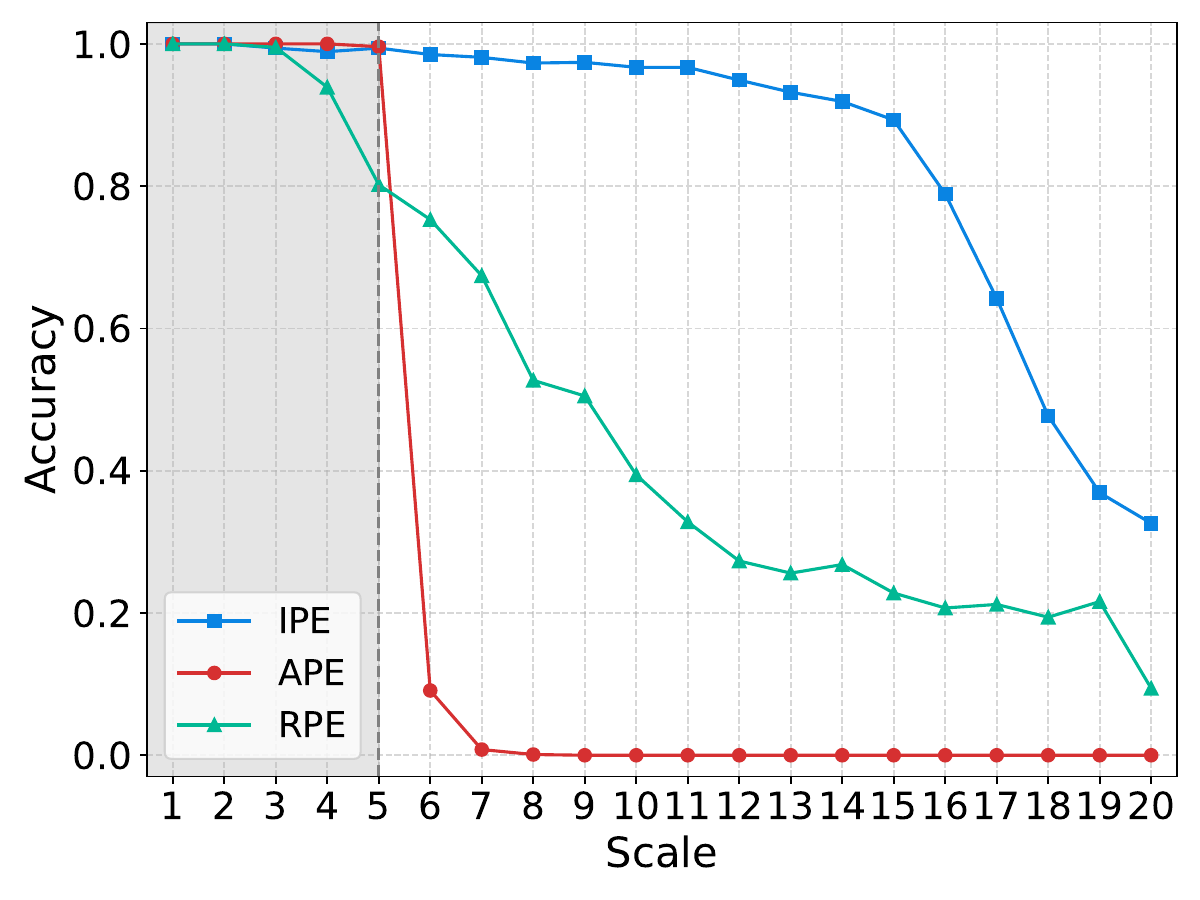}
    }%

    \caption{Evaluation results of models using different PEs across six tasks. Each model is trained on 10,000 samples of scales 1--5 for 300 epochs, with evaluation performed on 1,000 samples at each scale (1--20). Checkpoints are saved every 30 epochs. For each configuration, the plotted curve corresponds to the checkpoint that achieves the best average performance across all scales.}
    \label{fig:ipe-ape-rpe}
\end{figure}

We consider six tasks. For each task, the instances are all aligned to a target length to guarantee the existence of a characterizing PRF. Experimental results shown in Figure~\ref{fig:ipe-ape-rpe} demonstrate that when the operators required at larger scales are already present within the training domain, IPE outperforms both APE and RPE, especially in the three relatively complicated tasks: Addition, Multiplication, and Division. More details regarding experimental setups and implementations are provided in Appendix~\APPDXEXPERIMENTS 
.





\subsection{General PE Design Principles for LG}

\textbf{Characterizing PRF.} As our above analysis shows, the key to whether a PE can achieve LG for a task is whether the PRF of the PE can characterize some circuit representation of the task. Therefore, when designing a PE for a task, it is important to identify a PRF that properly characterizes the task and implement the PRF in the PE.

\textbf{Complexity-Generality Trade-Off.} Typically, there exist multiple circuit representations and thus various PRFs that characterize one task. On the one hand, choosing a PRF whose value set is smaller, corresponding to fewer or simpler operators, may have better LG performance or model efficiency. On the other hand, using a PRF that is too compact may limit the generality of the PE, making the design applicable to very restricted problems and sensitive to slight changes in the tasks. The extreme is IPE. IPE implements the characterizing PRF of the least value set. However, IPE for Copy aligned to scale 10 fails to achieve LG for Copy aligned to 20, just a subtle change in the data format. On the contrary, RPE encodes a PRF with a larger value set. While RPE is less effective compared to IPE in each task, it can be applied to Copy aligned both 10 and 20 without any modification.

\section{Extensions}\label{sec:extensions}

\begin{figure*}[th]
    \centering
    \subfloat[Addition, 5]{%
        \includegraphics[width=0.24\linewidth]{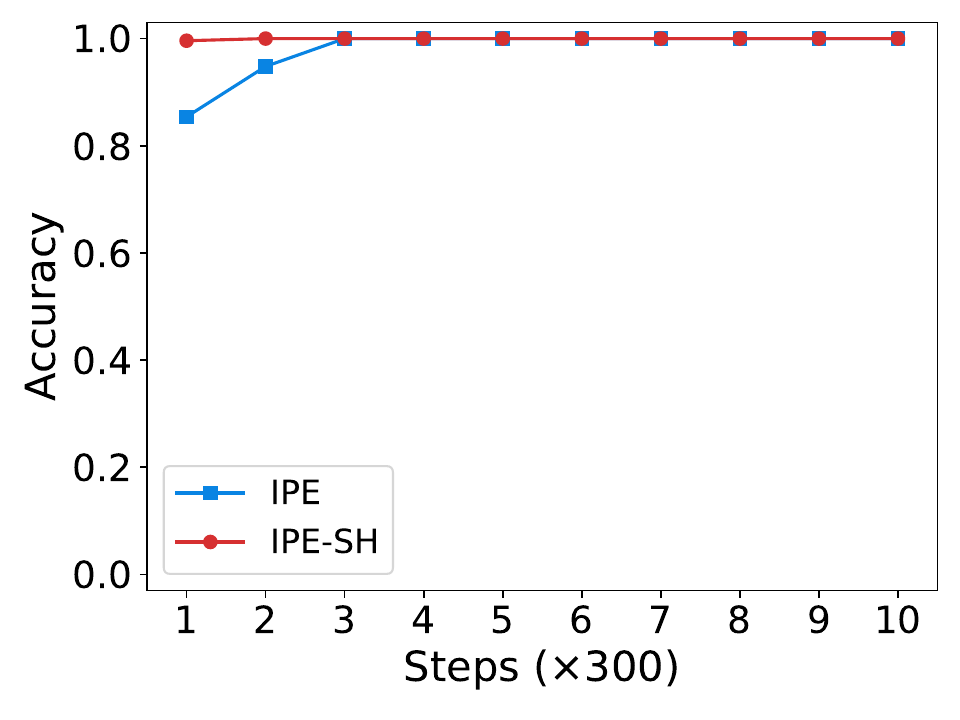}
    }%
    \hfill
    \subfloat[Addition, 10]{%
        \includegraphics[width=0.24\linewidth]{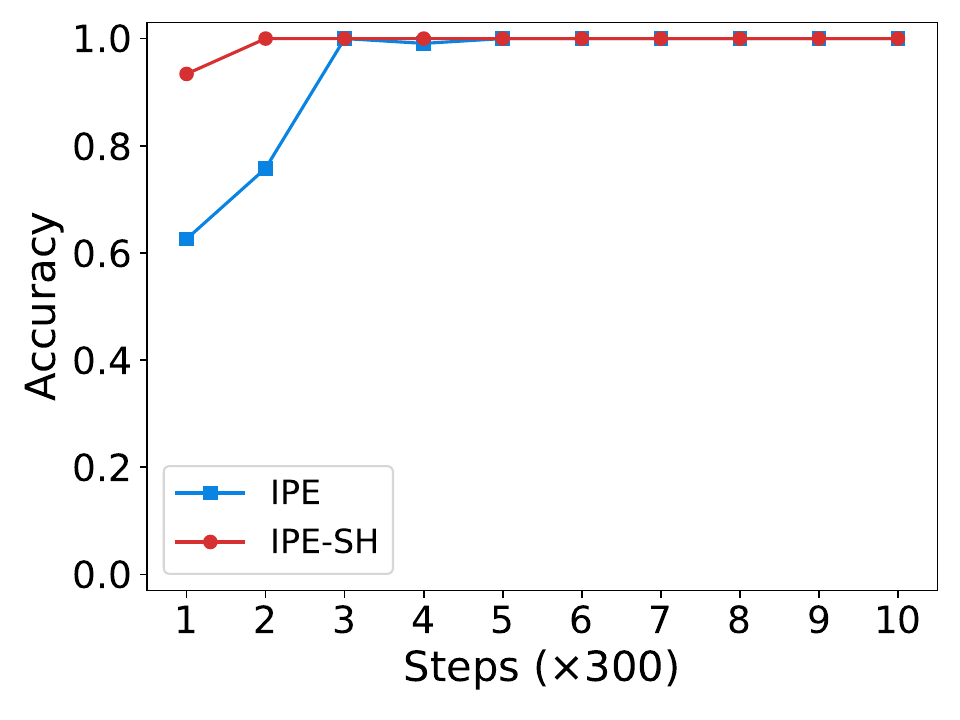}
    }%
    \hfill
    \subfloat[Addition, 15]{%
        \includegraphics[width=0.24\linewidth]{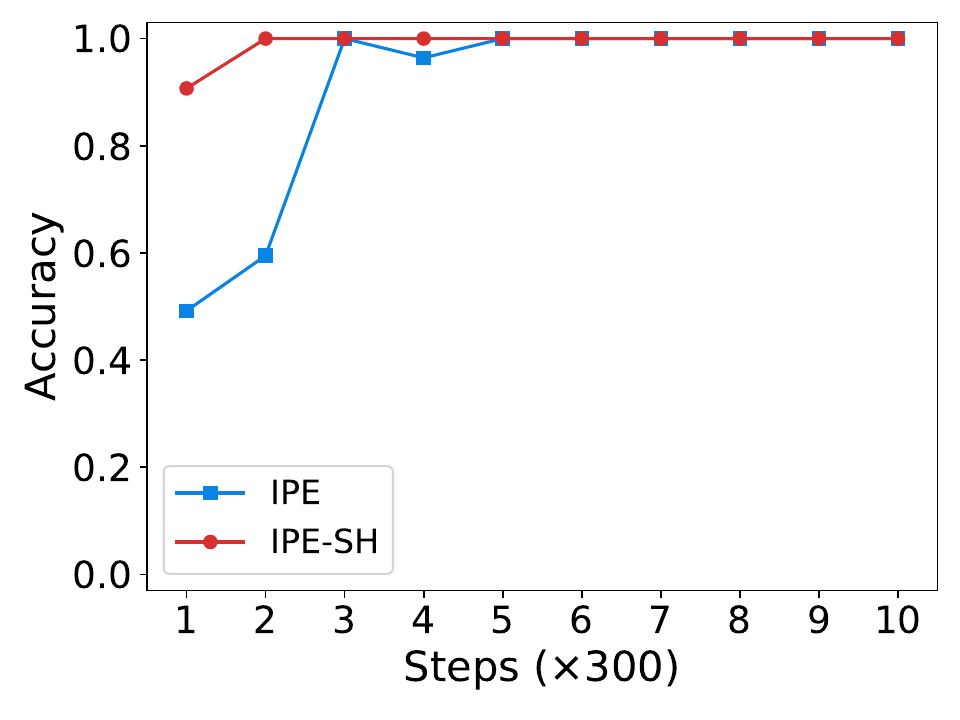}
    }%
    \hfill
    \subfloat[Addition, 20]{%
        \includegraphics[width=0.24\linewidth]{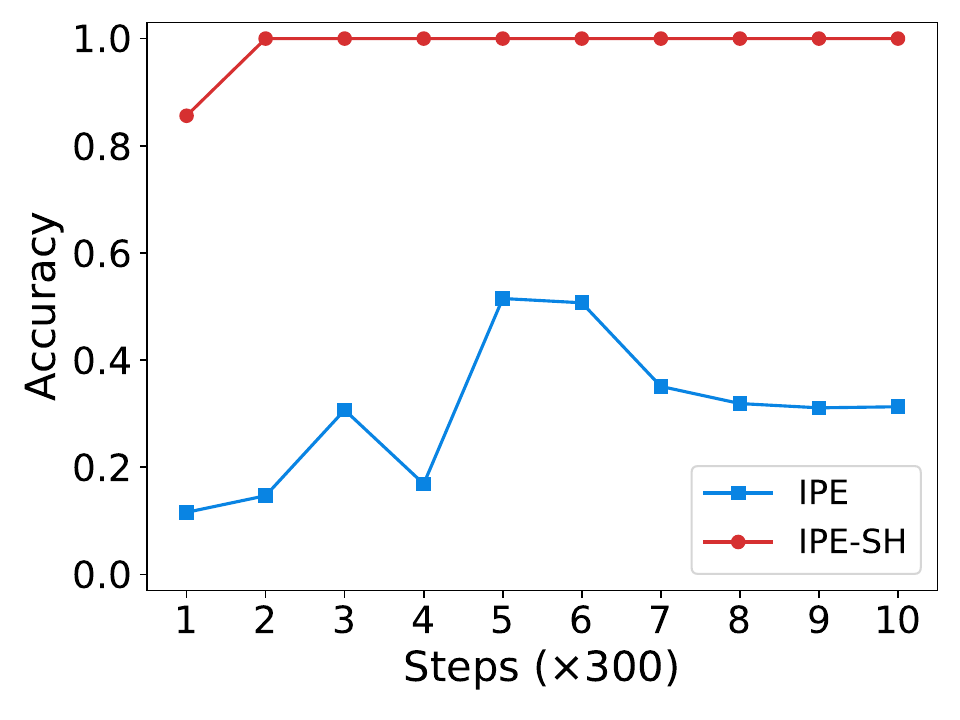}
    }%
    \hfill
    \subfloat[Multiplication, 5]{%
        \includegraphics[width=0.24\linewidth]{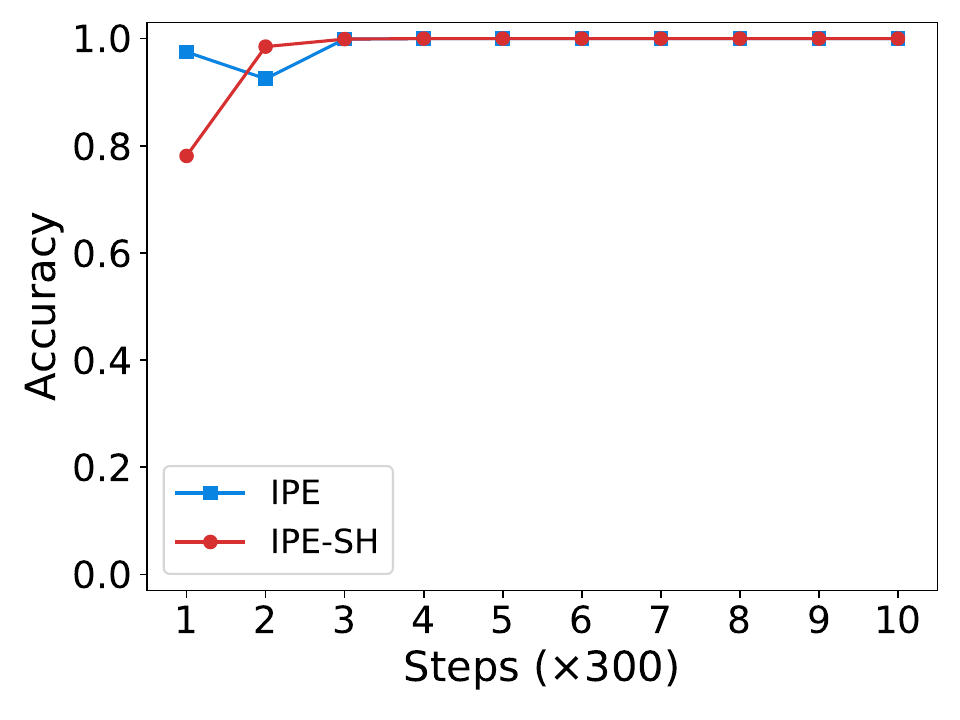}
    }%
    \hfill
    \subfloat[Multiplication, 10]{%
        \includegraphics[width=0.24\linewidth]{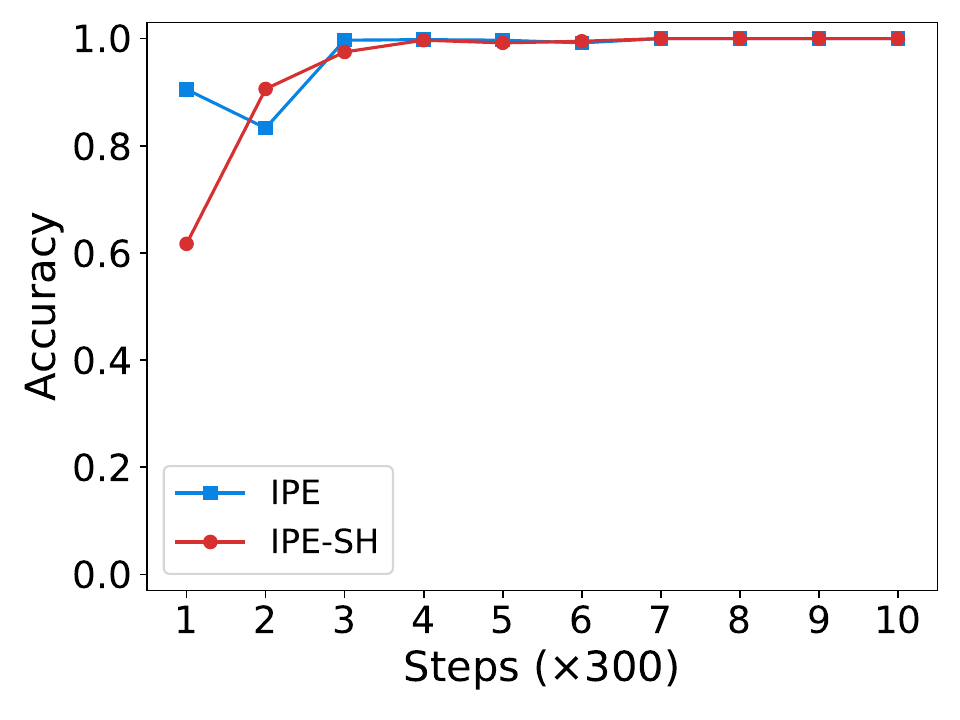}
    }%
    \hfill
    \subfloat[Multiplication, 15]{%
        \includegraphics[width=0.24\linewidth]{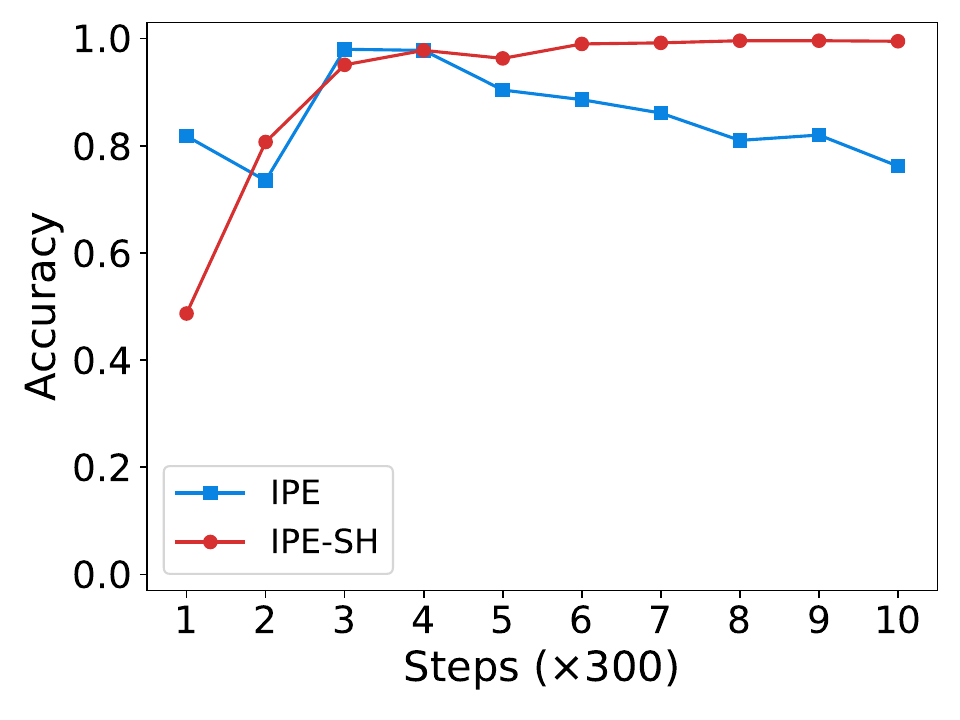}
    }%
    \hfill
    \subfloat[Multiplication, 20]{%
        \includegraphics[width=0.24\linewidth]{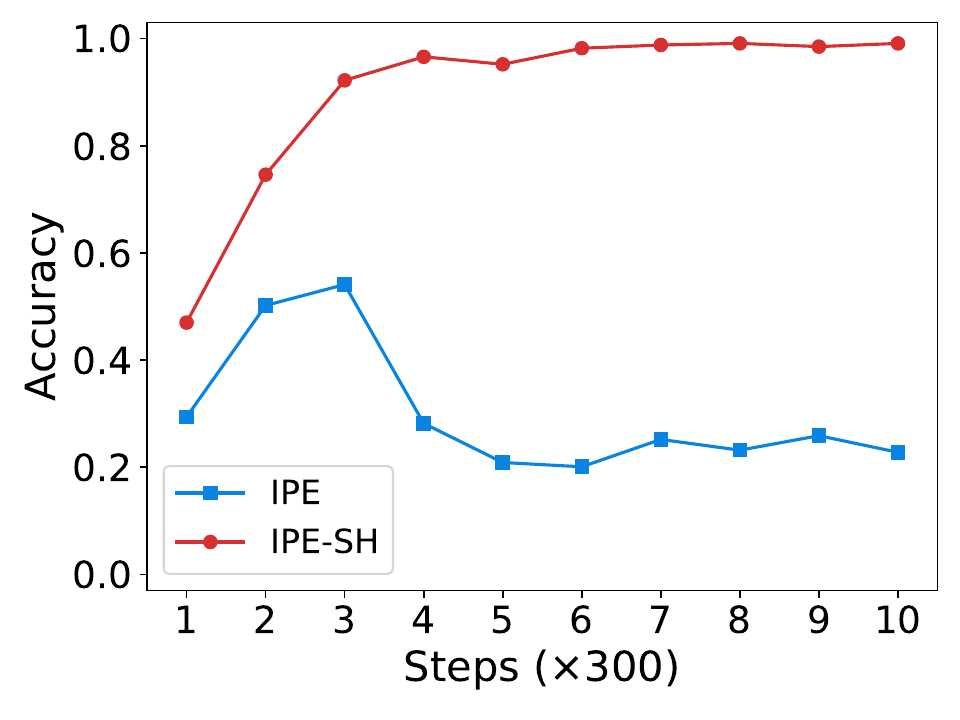}
    }%
    \hfill
    \subfloat[Division, 5]{%
        \includegraphics[width=0.24\linewidth]{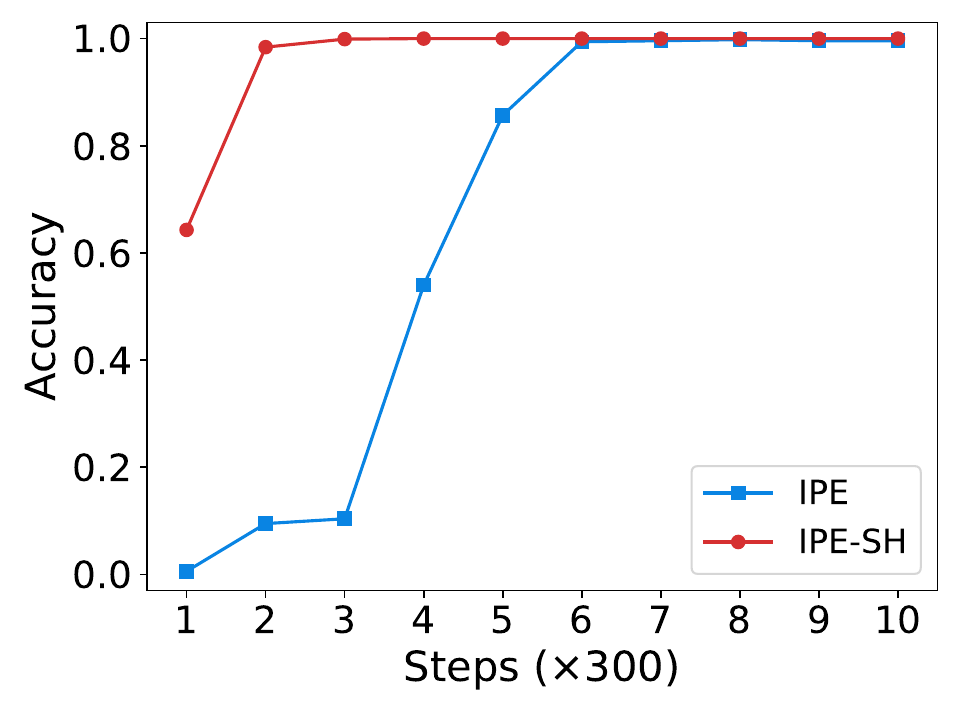}
    }%
    \hfill
    \subfloat[Division, 10]{%
        \includegraphics[width=0.24\linewidth]{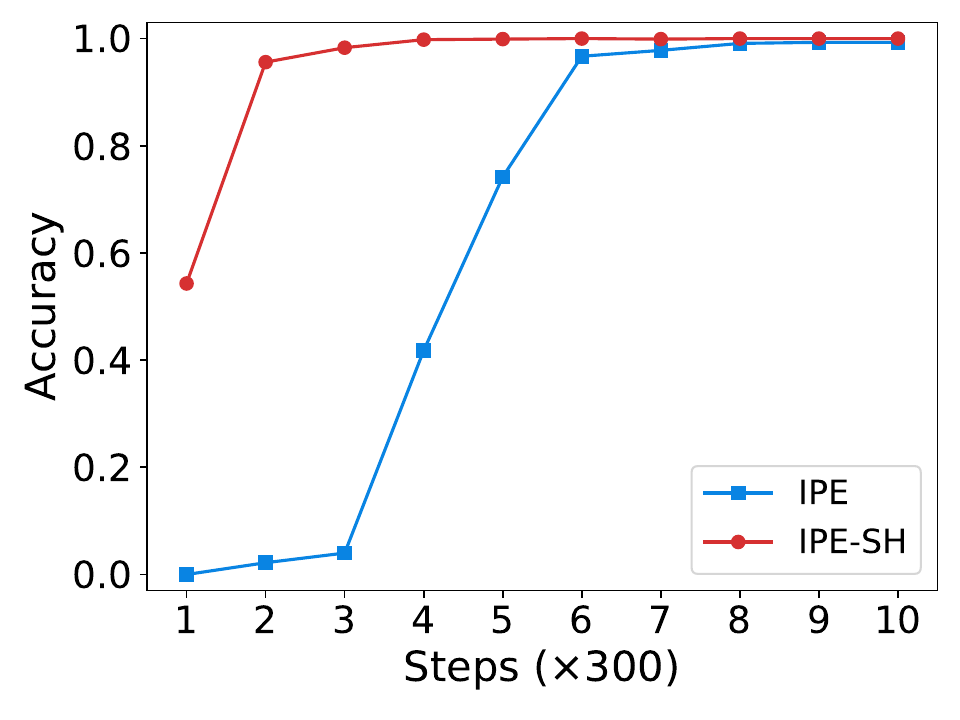}
    }%
    \hfill
    \subfloat[Division, 15]{%
        \includegraphics[width=0.24\linewidth]{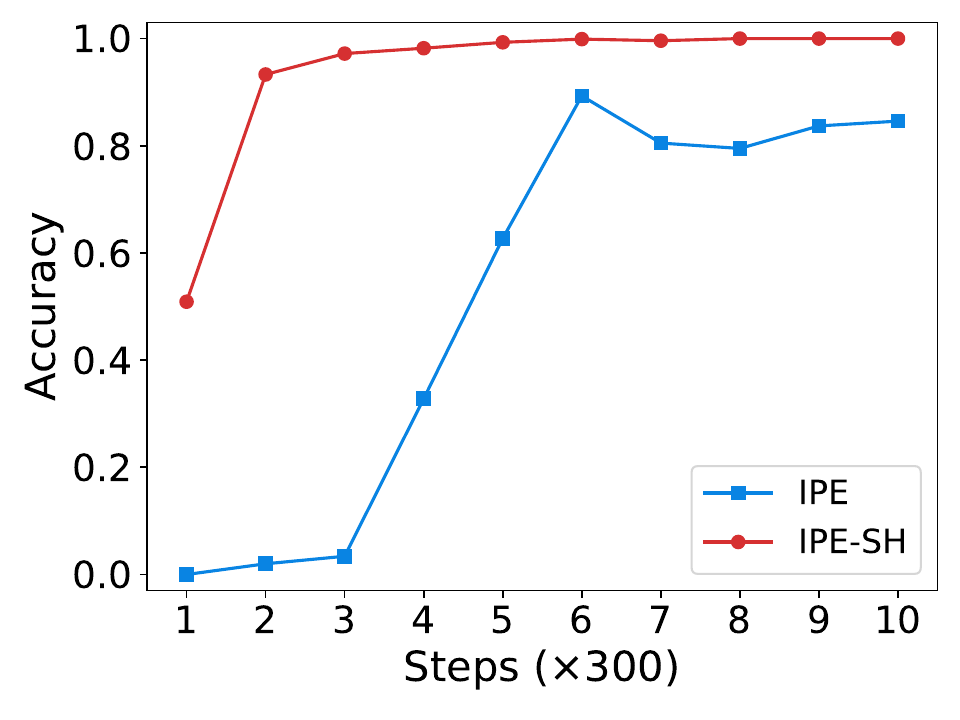}
    }%
    \hfill
    \subfloat[Division, 20]{%
        \includegraphics[width=0.24\linewidth]{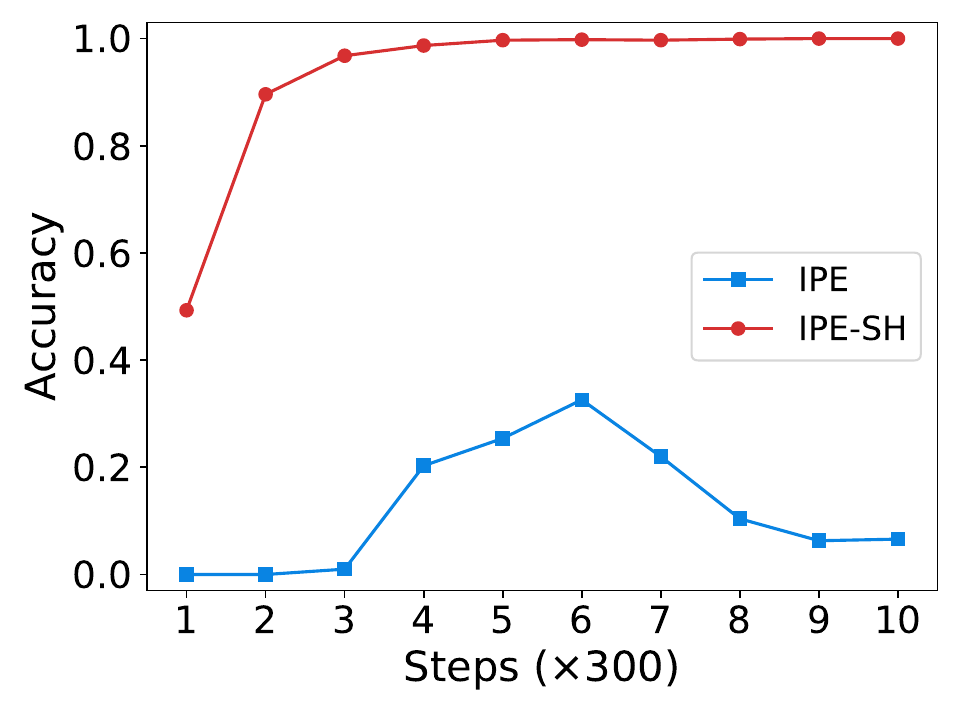}
    }%
    \caption{Comparison between IPE and IPE-SH in Addition, Multiplication (1 * N), and Division (N / 1). For IPE, we align input samples to scale 20, whereas IPE-SH operates without scale alignment. Both models are trained on samples of scales 1–5 and evaluated on scales 5, 10, 15, 20 (the numbers in the subpations mean the evaluation scales). For clarity, we present only the evaluation results on scales 16--20.}
    \label{fig:scale-hint}
\end{figure*}

Our analysis in Section~\ref{sec:practical-tf} shows that, in general, PEs can achieve LG only for tasks with circuit representations of non-increasing SRC, provided we choose PEs whose PRFs characterize these circuit representations. However, two practical issues arise: (1) It may not always be possible to design a PRF characterizing a circuit representation with non-increasing SRC; (2) Handcrafting a task-specific PE for each new task is impractical. To mitigate these issues, we propose two practical extensions: the \emph{scale hint} technique and \emph{learning-based position embeddings}.

\subsection{Scale Hint Technique}\label{subsec:sh}

In Section~\ref{sec:practical-tf}, we established that LG is achievable when the task has a circuit representation whose SRC does not increase from training to testing scales, provided a suitable PRF characterizing the corresponding circuit. However, standard PRFs are typically scale-invariant; hence, for certain circuit representations of non-increasing SRC, it may not be possible to construct any PRF that characterizes them.

\begin{example}[Non-Existence of Characterizing PRF]
    Consider the Addition task where the numbers are only aligned scale-wisely, i.e., an instance of scale $n$ is like
    \begin{equation*}
        \mathtt{x_1\;\dots\;x_n\;+\;y_1\;\dots\;y_n\;=\;z_1\; \dots\; z_n\; z_{n+1}},
    \end{equation*}
    for $n =1,\dots, N$. For such an unaligned Addition task, no circuit representation $\cC=\{C_n\}$ can be characterized by a scale-invariant PRF $\phi(i,j)$.
\end{example}

In practice, the scale of an instance is often known or can be reasonably estimated. The core idea of the Scale Hint (SH) technique is to leverage this information by augmenting the Position Representation Function (PRF) with the instance scale as an additional input. Formally, we define the PRF with Scale Hint (PRF-SH) as a mapping that explicitly incorporates the instance scale, i.e., $\phi:[N]\times [N]\times [N]\mapsto [S]$ that maps the query position $i$, the key position $j$, and the instance scale $n$ to some value $\phi(i,j,n)$. The resulting position embedding is denoted as PE-SH. Analogous to Definition~\ref{def:prf-char}, we can define when a PRF-SH characterizes a circuit representation.

Incorporating scale hints makes the PRF strictly more expressive, enabling the characterization of a broader class of circuit representations. In fact, the following Theorem~\ref{thm:scale-hint} shows that PRF-SH is complete for characterizing circuit representations of non-increasing SRC.

\begin{theorem}\label{thm:scale-hint} 
For any circuit representation $\cC=\{C_n\}$ of non-increasing SRC, there exists a corresponding PRF-SH that characterizes it.
\end{theorem}

Using PE-SH not only broadens the applicability but also enables more compact and efficient representations, thus facilitating more effective learning. For example, consider the Addition task. If we do not employ the scale hint, we need to align all the instances to the maximum target length:
\begin{equation*}
    \mathtt{x_1 \dots x_n} \underbrace{\mathtt{0 \dots 0}}_{N-n} + \mathtt{y_1 \dots y_n} \underbrace{\mathtt{0 \dots 0}}_{N-n} = \mathtt{z_1 \dots z_n z_{n+1}} \underbrace{\mathtt{0 \dots 0}}_{N-n}.
\end{equation*}
These redundant padding zeros increase computation and memory usage and may confound the model’s ability to identify the correct positions and operators. 
By contrast, if we apply the scale hint, we can represent the instance as:
\begin{equation*}
    \mathtt{x_1\;\dots\;x_n\;+\;y_1\;\dots\;y_n\;=\;z_1\;\dots\;z_n\;z_{n+1}},
\end{equation*}
which significantly reduces overhead when $n \ll N$. 
Figure~\ref{fig:scale-hint} is a comparison between IPE and IPE-SH in Multiplication and Division, respectively. The results demonstrate that incorporating a scale hint accelerates convergence and improves LG performance. 
Moreover, since no target length is fixed, PE-SH enables more flexible length generalization, allowing extrapolation to larger-scale instances beyond the limits imposed by fixed-length alignment.

\begin{figure*}[th]
    \centering
    \subfloat[SelectFirst]{%
        \includegraphics[width=0.32\linewidth]{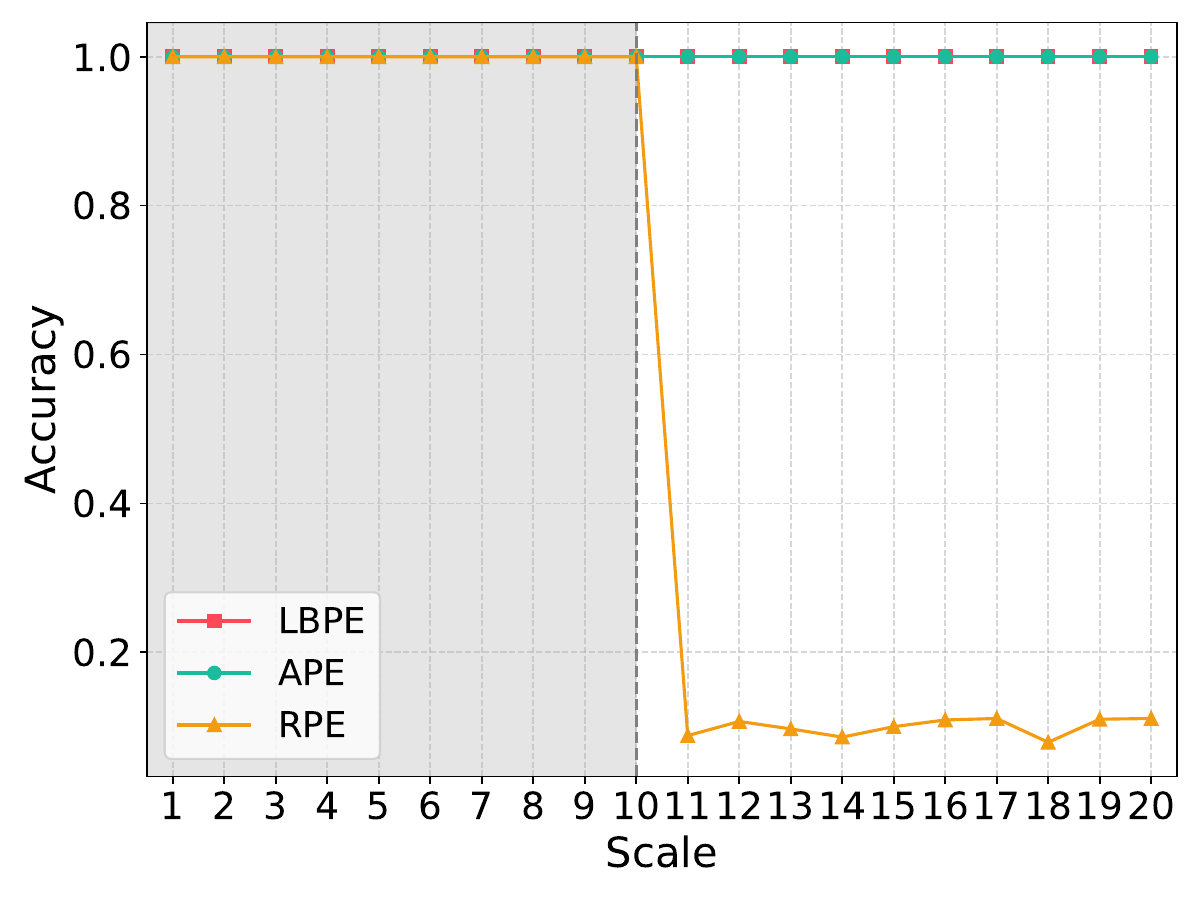}
    }%
    \hfill
    \subfloat[SelectMiddle]{%
        \includegraphics[width=0.32\linewidth]{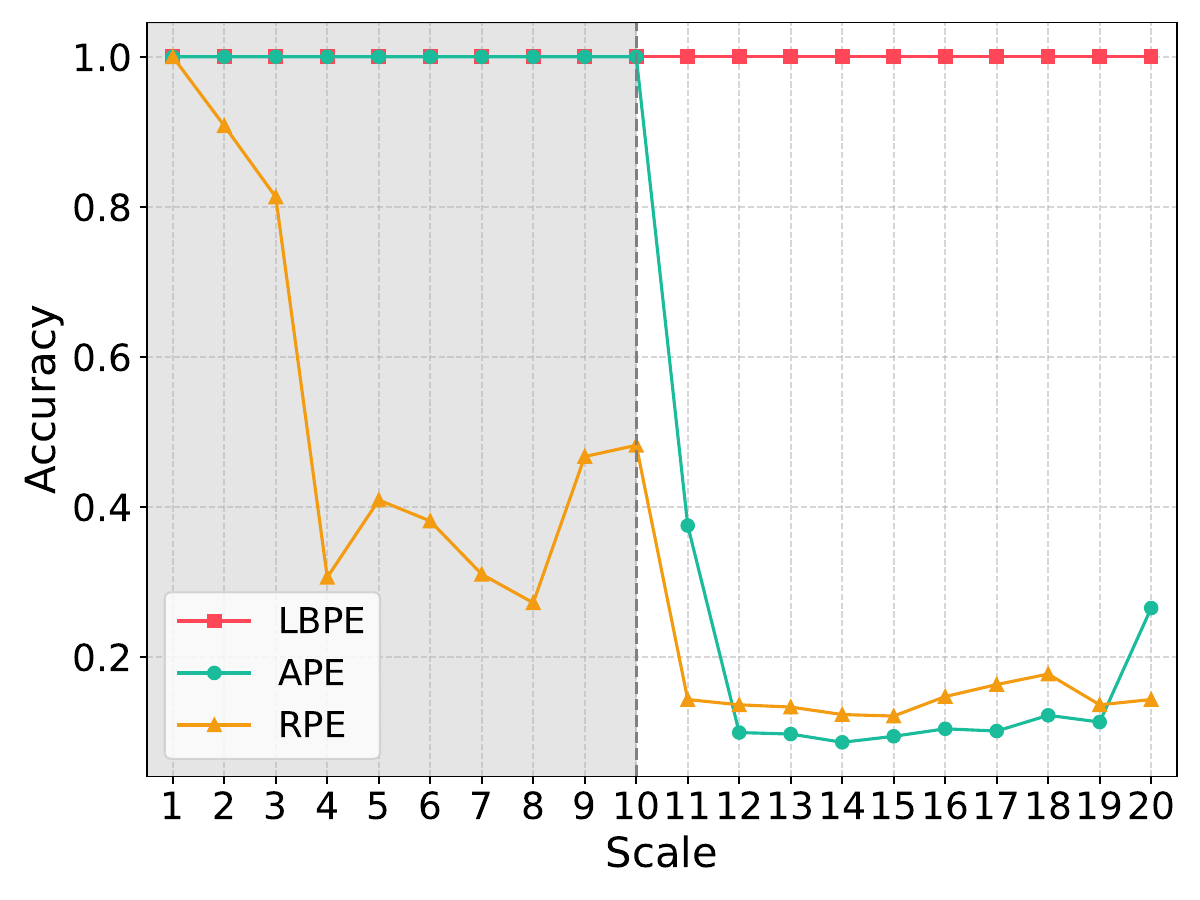}
    }%
    \hfill
    \subfloat[SelectLast]{%
        \includegraphics[width=0.32\linewidth]{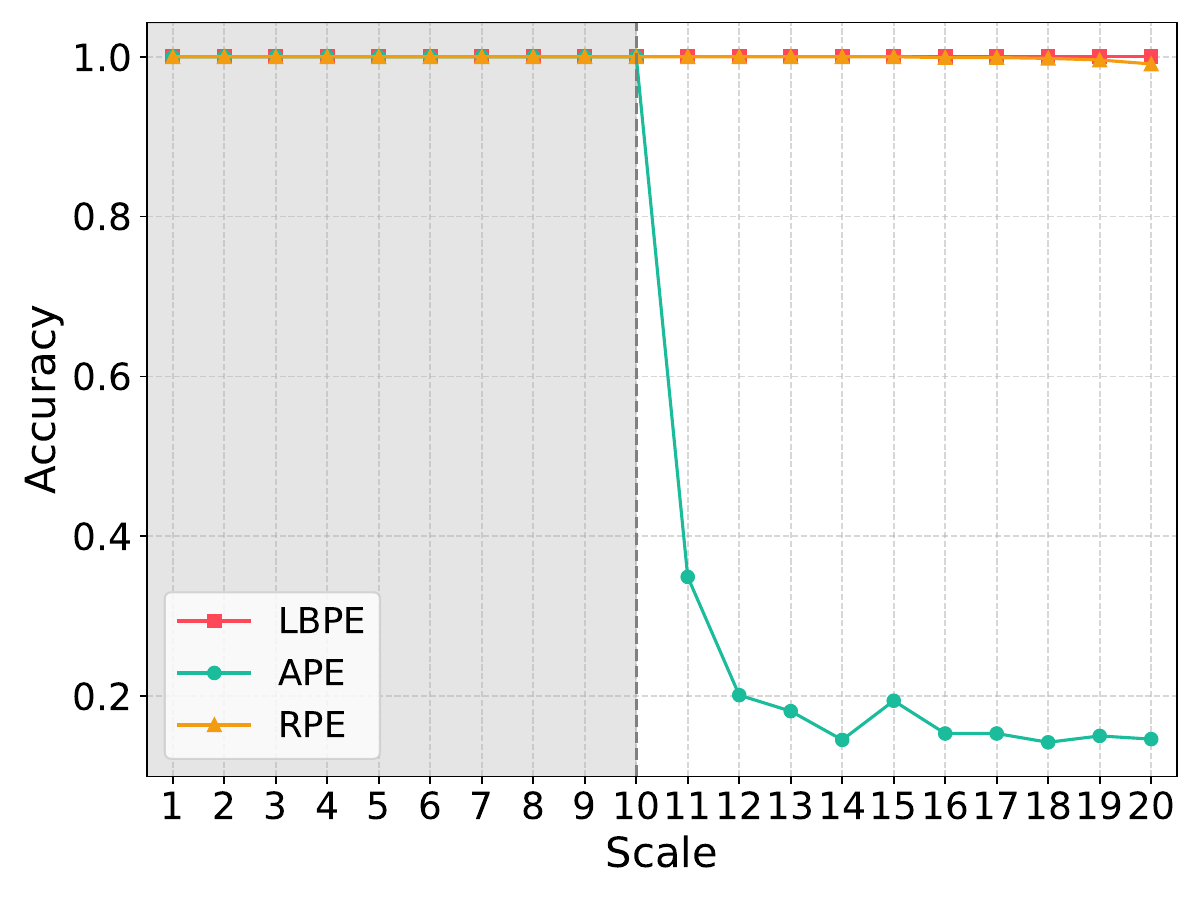}
    }%
    \caption{Evaluation results of models with LBPE across three different Select tasks (SelectFirst, SelectMiddle, and SelectLast). Each model is trained on 1,000 samples of scales 1--10 for 2,000 epochs and evaluated on 1,000 samples at each scale 1--20. We save checkpoints every 20 epochs. For each configuration, we plot the curve for the checkpoint of the best average performance across all scales.}
    \label{fig:lbpe}
\end{figure*}

\begin{figure}[th]
    \centering
    \includegraphics[width=0.95\linewidth]{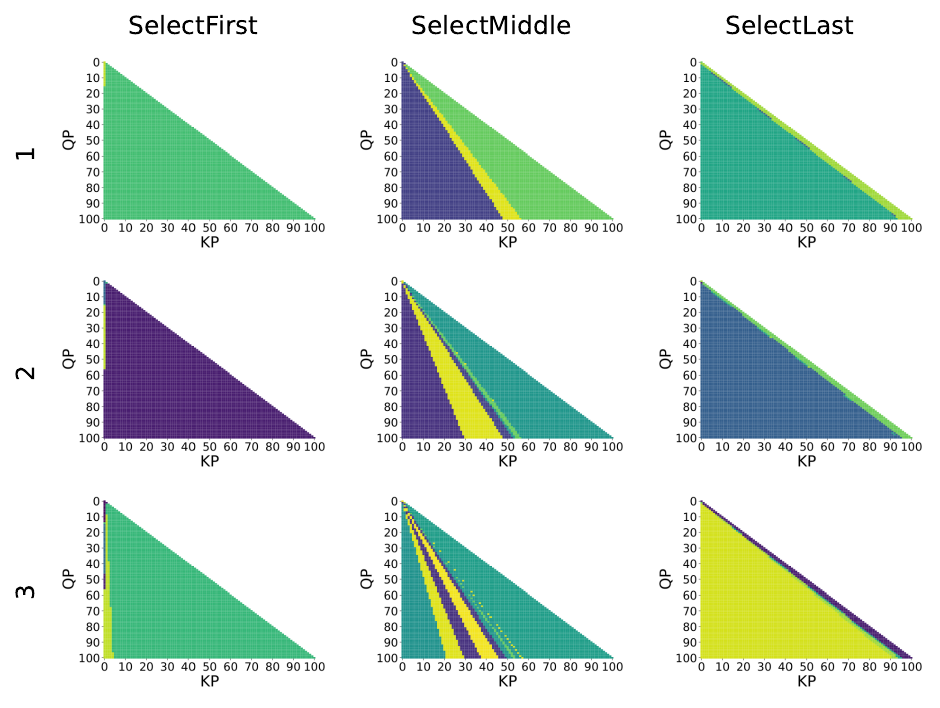}
    \caption{Visualization of the learned PRFs in the three Select tasks. For each task, we show the predicted PRF values corresponding to the top three prediction weights (ranked 1--3 from top to bottom) for each query position $i$ and key position $j$, where $i \leq j$. ``QP'' and ``KP'' mean ``query position'' and ``key position'', respectively.}
    \label{fig:lbprf}
\end{figure}

\subsection{Learning-Based Position Embeddings}\label{subsec:lbpe}

Thus far, we have demonstrated that handcrafting appropriate PEs (or PE-SH) enables LG whenever the SRC of the task does not increase from training to testing scales. However, practically speaking, perfect prior knowledge regarding the positional relationships within a task is usually unavailable. Moreover, designing new handcrafted PEs for each individual task is prohibitively expensive.

To address these limitations, we propose Learning-Based Position Embeddings (LBPE), in which the PRF (or PRF-SH) itself is made into a learnable component. Importantly, the proposed LBPE differs fundamentally from existing learnable PEs in prior literature. Specifically, the conventional learnable PE has a fixed PRF and learns embedding vectors, whereas our LBPE method explicitly learns the PRF function itself. 

LBPE can be implemented either with or without learnable embedding vectors. Here, we present one implementation of LBPE utilizing learnable embedding vectors for simplicity. Let $P\in\bbR^{S\times d}$ be the learnable embedding vectors and $\phi(i,j;\theta):[N]\times[N]\mapsto\Delta^{[S]}$ be the Learning-Based PRF (LBPRF), where $i, j$ are the query and key positions respectively, $S$ is an upper bound of the PRF value, and $\theta$ is the learnable parameter of the LBPRF block. Then the LBPE with the learnable parameters $\theta$ and $P$ is
\begin{equation*}
    \text{LBPE}(i,j;\theta, P) = P^\intercal \phi(i,j;\theta).
\end{equation*}
For notation simplicity, we write $\text{LBPE}(i,j;\theta, P)$ as $\text{LBPE}(i,j)$ when this does not lead to misunderstanding.

We can simply replace any learnable PE with the above $\text{LBPE}$. For instance, we can adapt a Transformer with key-only RPE (i.e., the RPE is only added to the key embedding) to use LBPE by replacing $\RPE_{i-j}$ with $\text{LBPE}(i,j)$. Denote the learnable $\RPE$ at layer $l$ by $\RPE_{i-j}^{(l)}$. The query-key weight $\alpha_{i,j}^{(l)}$ at layer $l$ from
\begin{equation*}
    \alpha_{i,j}^{(l)} = \left(h_j^{(l-1)} + \teal{\RPE}_{i-j}^{(l)}\right)^\intercal W_K^{(l)\intercal} W_Q^{(l)} h_i^{(l-1)},
\end{equation*}
becomes
\begin{equation*}
    \alpha_{i,j}^{(l)} = \left(h_j^{(l-1)} + \orange{\text{LBPE}}(i,j;\theta^{(l)},P^{(l)})\right)^\intercal W_K^{(l)\intercal} W_Q^{(l)} h_i^{(l-1)}.
\end{equation*}

However, we emphasize that LBPE is not restricted to this particular approach; other forms of PE such as RoPE can also be integrated into LBPE frameworks (see Appendix~\APPDXPEIMP 
).

We evaluate LBPE, implemented with learnable positional encodings, on three Select tasks that involve identifying tokens at specific positions. Specifically, we consider SelectFirst, SelectMiddle, and SelectLast, which correspond to selecting $x_1$, $x_{\lfloor n/2\rfloor + 1}$, and $x_n$, respectively, given an input of the form ``$x_1 \dots x_n =$.'' The results, presented in Figure~\ref{fig:lbprf}, show that APE and RPE achieve LG only in \textsc{SelectFirst} and \textsc{SelectLast}, respectively, whereas LBPE succeeds in all three tasks. Figure~\ref{fig:lbprf} also visualizes the learned PRFs, revealing that LBPE adaptively captures task-specific positional relationships that closely align with the characterizing PRFs of each task.

Similarly, we can directly combine LBPE with the SH technique, resulting in LBPE-SH:
\begin{equation*}
    \text{LBPE-SH}(i,j,n;\theta, P) = P^\intercal \phi(i,j,n;\theta).
\end{equation*}

\begin{figure}[th]
    \centering
    \subfloat[Copy]{%
        \includegraphics[width=0.48\linewidth]{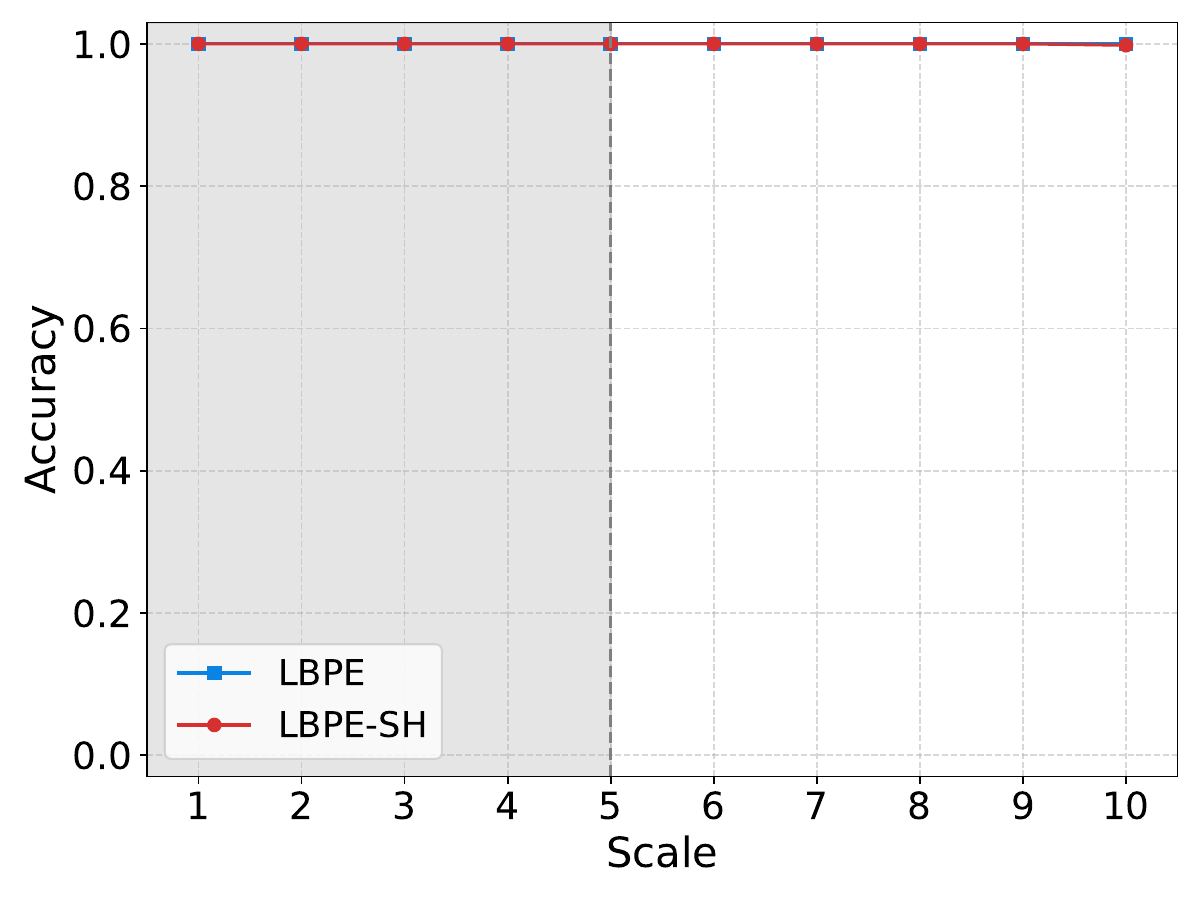}
    }%
    \hfill
    \subfloat[Reverse]{%
        \includegraphics[width=0.48\linewidth]{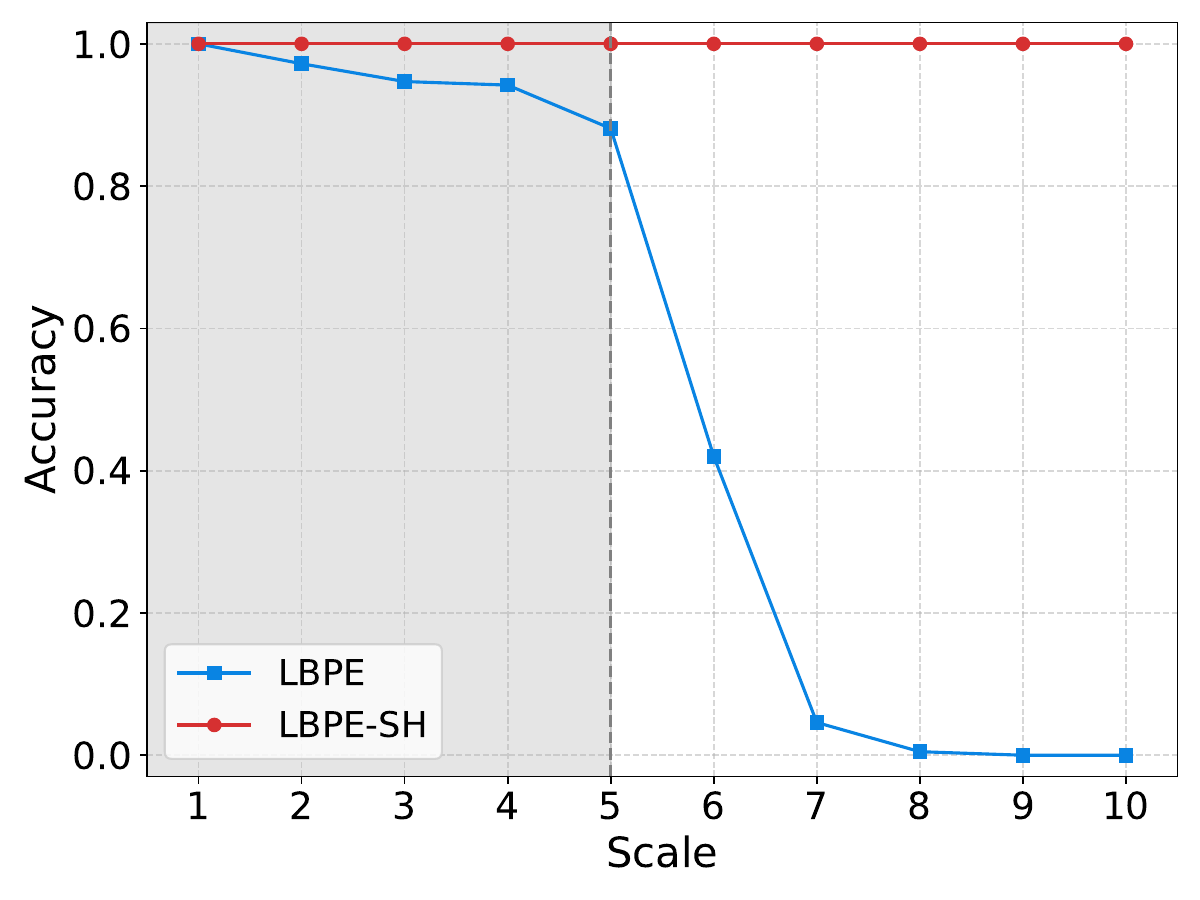}
    }%
    \\
    \subfloat[Shift]{%
        \includegraphics[width=0.48\linewidth]{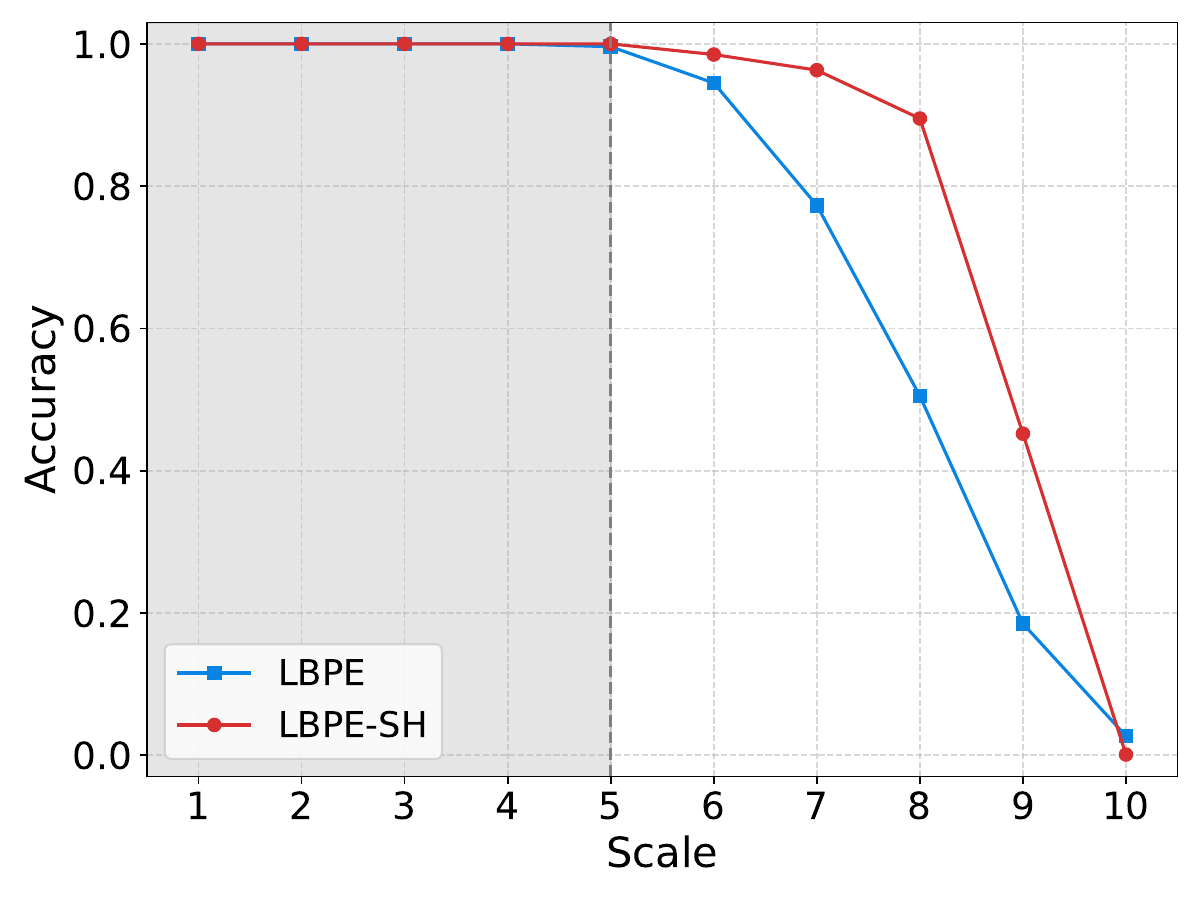}
    }%
    \hfill
    \subfloat[Parity (with CoT)]{%
        \includegraphics[width=0.48\linewidth]{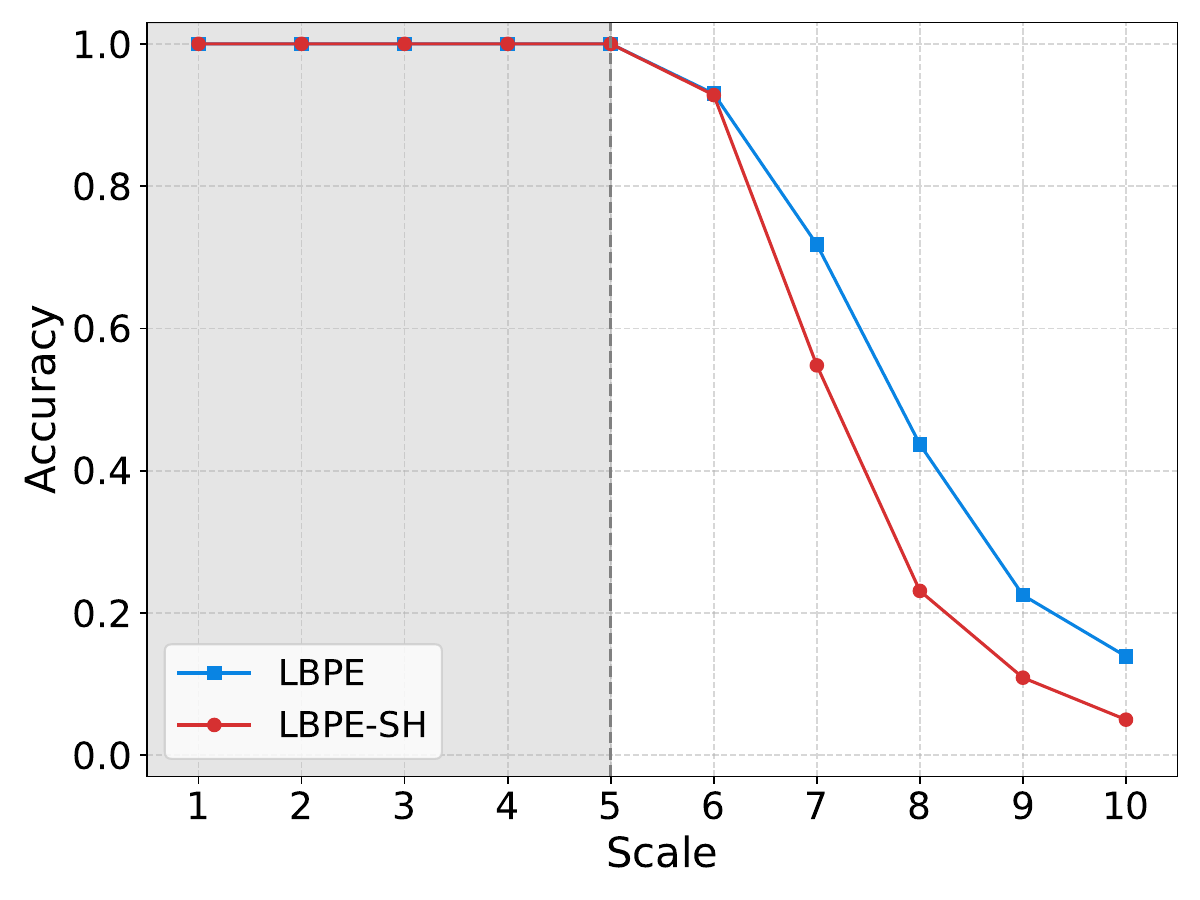}
    }%
    \\
    \subfloat[Multiplication (1 * N)]{%
        \includegraphics[width=0.48\linewidth]{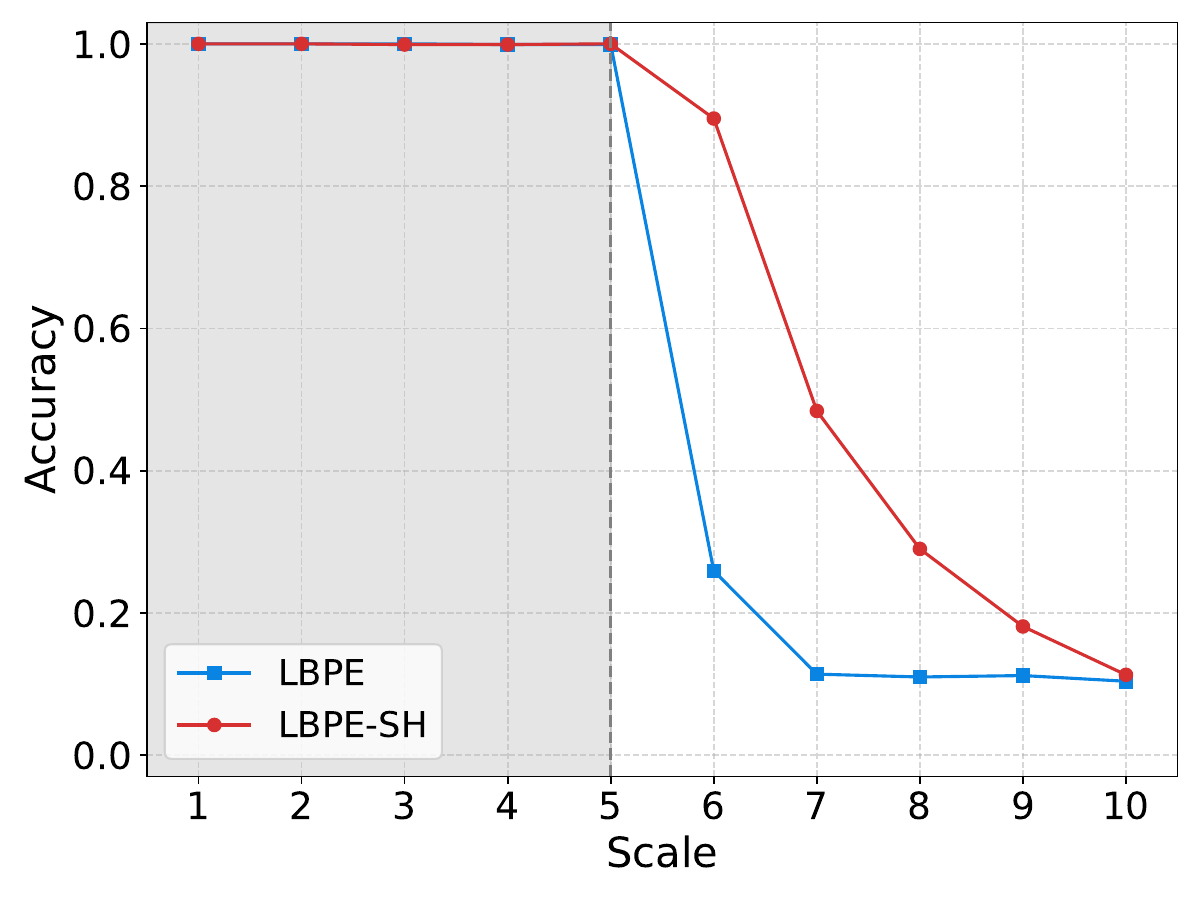}
    }%
    \hfill
    \subfloat[Division (N / 1)]{%
        \includegraphics[width=0.48\linewidth]{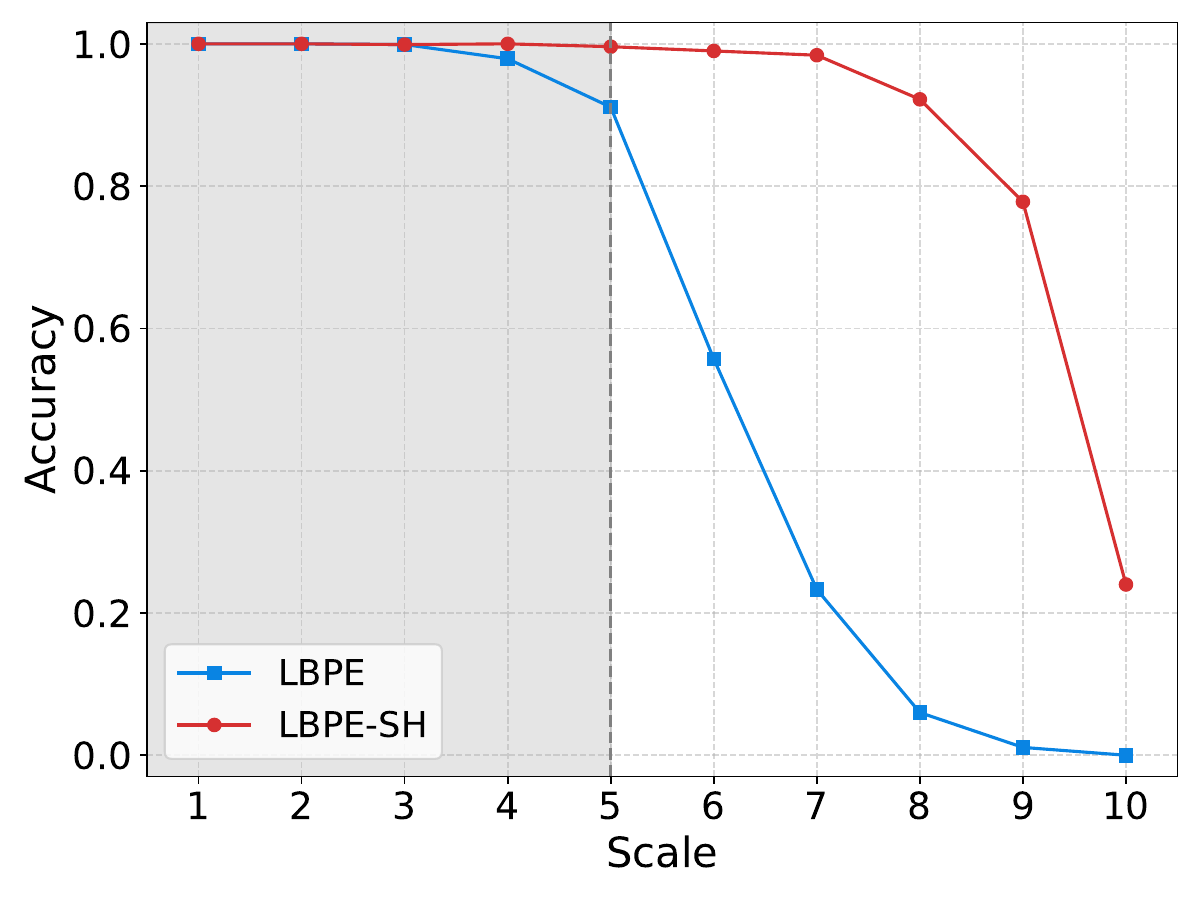}
    }%

    \caption{Evaluation results of models with LBPE and LBPE-SH on various tasks. Each model is trained on 10,000 samples from scales 1--5 for 10,000 epochs (equivalent to 100,000 steps under our hyperparameter setting). Checkpoints are saved every 10,00 steps and evaluated on 1,000 samples at each scale from 1 to 10. The curves are the evaluation results of the checkpoints achieving the best average accuracies (over all scales).}
    \label{fig:lbpe-lbpesh}
\end{figure}

For learning-based PEs, incorporating the SH technique can also enhance LG, similar to the effect observed with handcrafted PEs. As shown in Figure~\ref{fig:lbpe-lbpesh}, while both LBPE and LBPE-SH achieve LG in the Copy task, LBPE-SH outperforms LBPE in all other tasks except Parity.

Although LBPE offers the flexibility to automatically learn diverse positional relationships, it is unrealistic to expect a single LBPE model to achieve LG universally across all tasks. Task-specific prior knowledge remains essential for guiding the design of learning modules, as different architectural choices inherently bias the model toward capturing particular types of positional relationships. We leave a systematic investigation of how different architectures induce distinct biases in LBPEs as an important direction for future work.

\section{Conclusion}\label{sec:conclusion}

We analyze the role of PEs in LG. On the negative side, PEs have a fundamental limitation: they cannot facilitate the acquisition of new operators beyond those seen during training. On the positive side, PEs can align positions across different scales, enabling the model to apply learned operators to longer sequences. To effectively leverage PEs for LG, one needs to choose a PE with a characterizing PRF and strike a balance between its complexity and generality. We also propose the scale hint technique, extending the applicability of PEs to a wider range of tasks, and LBPE, alleviating the need for handcrafted PE design per task.

\textbf{Future Work.}
We only prove PE efficacy in POLA models. Rigorous theoretical analysis of the capabilities of PEs in practical Transformer architectures remains open for future work.
While we prioritize PRF analysis, the impact of different PE implementations warrants further study. Adapting the scale hint technique to natural language tasks where scales are less explicit and investigating how architectures modulate learned PRFs in LBPEs are also interesting future directions.

\bibliographystyle{IEEEtran}
\bibliography{references}


\newpage
\appendices
\onecolumn
\section{Hausdorff Measure and Hausdorff Dimension}\label{appdx:hausdorff}

To make our theoretical discussions self-contained, we briefly review the definitions and basic properties of Hausdorff measure and Hausdorff dimension. These concepts provide finer notions of ``size'' than Lebesgue outer measure, especially for sets of measure zero. The definitions and the properties that are useful for our proofs listed below follow standard references in fractal geometry; for a comprehensive treatment, we refer the reader to \cite{falconer2013fractal}.

\begin{definition}[Hausdorff Measure]
    Suppose that $F$ is a subset of $\bbR^n$ and $d$ is a non-negative number. For any $\delta > 0$, we define
    \begin{equation*}
        H_{\delta}^d(F)=\inf\left\{\sum_{i=1}^{\infty} \op{diam}(U_i)^d\,\middle|\, F\subseteq\sum_{i=1}^{\infty} U_i, \op{diam}(U_i)\leq\delta\right\},
    \end{equation*}
    where $\op{diam}(U):=\sup\left\{|x-y|\mid x,y\in U\right\}$ is the diameter of $U$. The $d$-dimensional Hausdorff measure of $F$ is defined as
    \begin{equation*}
        H^d(F) := \lim_{\delta\to 0} H_\delta^d (F).
    \end{equation*}
\end{definition}

\begin{definition}[Hausdorff Dimension]
    Suppose that $F$ is a subset of $\bbR^n$. The Hausdorff dimension of $F$ is defined as 
    \begin{equation*}
        \dim_{H}(F) 
        = \inf\left\{d\ge 0\mid H^d(F)=0\right\}
        = \sup\left\{d\mid H^d(F)=\infty\right\}.
    \end{equation*}
\end{definition}

\begin{proposition}
    Suppose that $F$ is a subset of $\bbR^n$ and $d$ is a non-negative number. The following properties hold:
    \begin{itemize}
        \item (Monotonicity) If $E\subseteq F$, then $H^d(E)\leq H^d(F)$ and $\dim_H(E)\leq\dim_H(F)$.
        \item (Countable Subadditivity) If $F=\bigcup_{i=1}^{\infty} F_i$, then $H^d(F)\leq \sum_{i=1}^\infty H^d(F_i)$.
        \item (Countable Stability) If $F=\bigcup_{i=1}^{\infty} F_i$, then $\dim_H(F)=\sup_{1\leq i<\infty}\{dim_H(F_i)\}$.
        \item (Countable sets) If $F$ is countable, then $\dim_H(F)=0$.
    \end{itemize}
\end{proposition}

Intuitively, Hausdorff measure generalizes familiar notions such as length, area, and volume to arbitrary real dimensions. Hausdorff dimension of a set is the critical threshold at which this measure drops from infinity to zero, quantifying how ``dense'' or ``complex'' the set is at arbitrarily small scales.
\section{No-Free-Lunch Theorem of LG}\label{appdx:nfl_lg}

No-Free-Lunch Theorem of LG states that the average performance of any two learning algorithms over all possible target concepts is identical. 

\begin{theorem}[No-Free-Lunch Theorem of LG \cite{chen2025low}]\label{thm:nfl-lg}
    For some $N > N_0$, consider two sets $\cX_{N_0}=\Sigma^{[N_0]}$ and $\cX_N=\Sigma^{[N]}$. Let $\cY$ be a finite set.
    Let $c_1,c_2\in\cF(:=\cF_{\cX_N, \cY})$ be two concepts such that $c_1(x)=c_2(x)$ for all $x\in\cX_{N_0}$. 
    For any $c\in\cF$ and $\cX'\subseteq\cX$, define $\cF/\left(c\mid\cX'\right):=\left\{f\in\cF\mid f(x)=c(x)\text{ for all } x\in\cX'\right\}$. Let $\ell:\cY\times\cY\mapsto\bbR$ be the loss function. For any distribution $\cD(\cX_{N})$ such that $\supp\left(\cD(\cX_N)\right)=\cX_N$:
    \begin{equation*}
        \sum_{f\in\cF/\left(c_1|_{\cX_{N_0}}\right)} \E_{x\sim \cD(\cX_N)}\left[\ell\left(c_1(x), f(x)\right)\right] = 
        \sum_{f\in\cF/\left(c_2|_{\cX_{N_0}}\right)} \E_{x\sim \cD(\cX_N)}\left[\ell\left(c_2(x), f(x)\right)\right].
    \end{equation*}
\end{theorem}

Theorem~\ref{thm:nfl-lg} implies that no learning algorithm can universally achieve length generalization across all possible target functions. Consequently, incorporating prior knowledge about the target becomes essential for selecting algorithms or designing models capable of LG.

We note that the negative results, i.e., the limitations of PEs, are independent of the above No-Free-Lunch Theorem. We show that when shifting from short to long instances requires ``new operators'', LG cannot be achieved solely by adapting the PE, even under perfect prior knowledge.
\section{Different Implementations of PEs}\label{appdx:pe-imp}

PEs may share the same PRFs but differ in their concrete implementations. 
In the following subsections, we present distinct implementations of the same PRF $\phi:\bbN\times\bbN\mapsto\bbN$. Without loss of generality, we consider sequences up to length $N$ and the PRF $\phi([N]\times[N])\subseteq [S]$. We denote the input sequence by $\bm{h}^{(l)}=\left(h_1^{(l)},\dots, h_n^{(l)}\right)$ and the query-key weight by $\alpha_{i,j}^{(l)}$ for layer $l$.

\subsection{Learnable PE}

In learnable PE \cite{shaw2018self}, we encode the PRF with learnable embedding vectors. Two pairs $(i,j)$ and $(i',j')$ are encoded with the same embedding vector if $\phi(i,j)=\phi(i',j')$. The PE vectors are typically added to the hidden states. Let $P\in\bbR^{S\times d}$ be the learnable embedding vectors. Then the query-key weight is computed as
\begin{equation*}
    \alpha_{i,j}^{(l)} = \left(h_j^{(l-1)} + P_{\phi(i,j)}^{(l)}\right)^\intercal W_K^{(l)\intercal} W_Q^{(l)} h_i^{(l-1)}.
\end{equation*}

The above implementation adds the PE vectors to the hidden states only when computing the keys. We can also add the PE vectors when computing the queries and the values, though these may have little effect on the performance \cite{shaw2018self}. 

\subsection{Rotary PE}

Rotary PE (RoPE) \cite{su2024roformer} uses multiplication instead of addition to incorporate positional information. For a sequence $\bm x = \left(x_1, \dots, x_n\right)$, it seeks embedding functions $f_Q, f_K:\bbR^d\times\mapsto [N]$ such that 
\begin{equation*}
    \left\langle f_Q(x_i, i), f_K(x_j, j)\right\rangle
    =
    g\left(x_i, x_j, \phi(i,j)\right),
\end{equation*}
for some function $g:\bbR^d\times\bbR^d\times [S]\mapsto\bbR$.

If there exist two functions $\phi_1:[N]\mapsto\bbN$ and $\phi_2:[N]\mapsto\bbN$ such that $\phi(i,j)=\phi_1(i)-\phi_2(j)$, then we can set
\begin{equation*}
    \begin{gathered}
        f_Q(x_i, i) = R^{Q,d}_{\Theta, i} W_Q x_i, \\
        f_K(x_j, j) = R^{K,d}_{\Theta, j} W_K x_j,
    \end{gathered}
\end{equation*}
where
\begin{equation*}
    \begin{gathered}
        R^{Q,d}_{\Theta,i} =
        \begin{bmatrix}
            \cos \phi_1(i) \theta_1 & -\sin \phi_1(i)\theta_1 & \cdots & 0 & 0 \\
            \sin \phi_1(i) \theta_1 & \cos \phi_1(i)\theta_1 & \cdots & 0 & 0 \\
            \vdots & \vdots & \ddots & \vdots & \vdots \\
            0 & 0 & \cdots & \cos \phi_1(i) \theta_{d/2} & -\sin\phi_1(i)\theta_{d/2}\\
            0 & 0 & \cdots & \sin \phi_1(i) \theta_{d/2} & \cos \phi_1(i)\theta_{d/2}
        \end{bmatrix},\\
        R^{K,d}_{\Theta,j} =
        \begin{bmatrix}
            \cos \phi_2(j) \theta_1 & -\sin \phi_2(j)\theta_1 & \cdots & 0 & 0 \\
            \sin \phi_2(j) \theta_1 & \cos \phi_2(j)\theta_1 & \cdots & 0 & 0 \\
            \vdots & \vdots & \ddots & \vdots & \vdots \\
            0 & 0 & \cdots & \cos \phi_2(j) \theta_{d/2} & -\sin \phi_2(j)\theta_{d/2}\\
            0 & 0 & \cdots & \sin \phi_2(j) \theta_{d/2} & \cos \phi_2(j)\theta_{d/2}
        \end{bmatrix},
    \end{gathered}
\end{equation*}
with the hyperparameter $\Theta=\{\theta_1,\dots,\theta_{d/2}\}$.

If the PRF $\phi(i,j)$ cannot be decomposed as the different of $\phi_1(i)$ and $\phi_2(j)$, we can compute $g\left(x_i,x_j,\phi(i,j)\right)$ directly:
\begin{equation*}
    g\left(x_i,x_j,\phi(i,j)\right) = \left(W_Q x_i\right)^\intercal
    R^{Q,K,d}_{\Theta,i,j} \left(W_K x_j\right),
\end{equation*}
where
\begin{equation*}
        R^{Q,K,d}_{\Theta,i,j} =
        \begin{bmatrix}
            \cos \phi(i,j) \theta_1 & -\sin \phi(i,j)\theta_1 & \cdots & 0 & 0 \\
            \sin \phi(i,j) \theta_1 & \cos \phi(i,j)\theta_1 & \cdots & 0 & 0 \\
            \vdots & \vdots & \ddots & \vdots & \vdots \\
            0 & 0 & \cdots & \cos \phi(i,j) \theta_{d/2} & -\sin \phi(i,j) \theta_{d/2}\\
            0 & 0 & \cdots & \sin \phi(i,j) \theta_{d/2} & \cos \phi(i,j) \theta_{d/2}
        \end{bmatrix}.
\end{equation*}

The query-key weight is computed as
\begin{equation*}
    \alpha_{i,j}^{(l)} = g\left(h^{(l-1)}_i,h^{(l-1)}_j,\phi(i,j)\right).
\end{equation*}

While LBPE is implemented with learnable PE in Section~\ref{sec:extensions}, it can also be combined with RoPE. If we have prior knowledge or a belief that the target PRF is decomposable, we consider two models $\tilde{\phi}_1(\cdot;w_1), \tilde{\phi_2}(\cdot;w_2):[N]\mapsto \Delta^{[S]}$, where $w_1,w_2$ are learnable parameters. We replace $R_{\Theta,i}^{Q,d}$ and $R_{\Theta,j}^{K,d}$ with their learning-based variants $\tilde{R}_{\Theta,i}^{Q,d}(w_1)$ and $\tilde{R}_{\Theta,j}^{K,d}(w_2)$, where
\begin{equation*}
    \begin{gathered}
        \tilde{R}^{Q,d}_{\Theta,i}(w_1) =
        \sum_{s\in[S]}\left[\tilde{\phi}_1(i;w_1)\right]_s
        \begin{bmatrix}
            \cos s \theta_1 & -\sin s\theta_1 & \cdots & 0 & 0 \\
            \sin s\theta_1 & \cos s\theta_1 & \cdots & 0 & 0 \\
            \vdots & \vdots & \ddots & \vdots & \vdots \\
            0 & 0 & \cdots & \cos s\theta_{d/2} & -\sin s\theta_{d/2}\\
            0 & 0 & \cdots & \sin s\theta_{d/2} & \cos s\theta_{d/2}
        \end{bmatrix},\\
        \tilde{R}^{K,d}_{\Theta,j}(w_2) =
        \sum_{s\in[S]}\left[\tilde{\phi}_2(j;w_2)\right]_s
        \begin{bmatrix}
            \cos s\theta_1 & -\sin s\theta_1 & \cdots & 0 & 0 \\
            \sin s\theta_1 & \cos s\theta_1 & \cdots & 0 & 0 \\
            \vdots & \vdots & \ddots & \vdots & \vdots \\
            0 & 0 & \cdots & \cos s\theta_{d/2} & -\sin s\theta_{d/2}\\
            0 & 0 & \cdots & \sin s\theta_{d/2} & \cos s\theta_{d/2}
        \end{bmatrix}.
    \end{gathered}
\end{equation*}

In the general case, we can replace $R^{Q,K,d}_{\Theta,i,j}$ with its learning-based counterpart $\tilde{R}^{Q,K,d}_{\Theta,i,j}(w)$ defined as
\begin{equation*}
    \tilde{R}^{Q,K,d}_{\Theta,i,j}(w) =
    \sum_{s\in[S]}\left[\tilde{\phi}(i,j;w)\right]_s
    \begin{bmatrix}
        \cos s \theta_1 & -\sin s\theta_1 & \cdots & 0 & 0 \\
        \sin s\theta_1 & \cos s\theta_1 & \cdots & 0 & 0 \\
        \vdots & \vdots & \ddots & \vdots & \vdots \\
        0 & 0 & \cdots & \cos s\theta_{d/2} & -\sin s\theta_{d/2}\\
        0 & 0 & \cdots & \sin s\theta_{d/2} & \cos s\theta_{d/2}
    \end{bmatrix},
\end{equation*}
where $\phi(\cdot,\cdot;w): [N]\times [N]\mapsto \Delta^{[S]}$ is the LBPRF with the learnable parameter $w$.

\section{Proofs}\label{appdx:proof}

\subsection{Proof for Theorem~\ref{thm:pola-limitation}}

\begin{lemma}\label{lm:hausdoff-dim}
    Let $A, B\subseteq \bbR^d$. If $\dim_H(B) < \dim_H(A)$, then
    \begin{equation*}
        \dim_H(A\setminus B) = \dim_H (A), \text{ and } 
        H^{\dim_H(A)}(A\setminus B) = H^{\dim_H(A)}(A).
    \end{equation*}
\end{lemma}

\begin{proof}[Proof for Lemma~\ref{lm:hausdoff-dim}]
    For notation simplicity, define $d_A:=\dim_H(A)$.
    Assume that $\dim_H(A\setminus B)<d_A$. Then there exists a $d'$ such that $\max\left\{\dim_H(A\setminus B),\dim_H(B)\right\}<d'<d_A$. By the countable subadditivity of the Hausdorff measure \cite{falconer2013fractal}, we have
    \begin{equation*}
        0\leq H^{d'}(A) \leq H^{d'}(A\setminus B) + H^{d'}(B) = 0 + 0 = 0.
    \end{equation*}
    Hence, we obtain that $H^{d'}(A)=0$, which contradicts the definition of the Hausdorff dimension $d_A$.

    It remains to prove that $H^{d_A}(A\setminus B) = H^{d_A}(A)$.
    Since $\dim_H(B) < \dim_H(A)$, we have $H^{d_A}(B)=0$. On the one hand, we have
    \begin{equation*}
        H^{d_A}(A) \leq H^{d_A}(A\setminus B) + H^{d_A}(B)=H^{d_A}(A\setminus B).
    \end{equation*}
    On the other hand, it holds that
    \begin{equation*}
        H^{d_A}(A) \ge H^{d_A}(A\setminus B).
    \end{equation*}
    Therefore, we have $H^{d_A}(A\setminus B) = H^{d_A}(A)$.
\end{proof}

Note that $d_0:=\dim_H\left(\cF_{M,B_0}\right)=\frac{N (N+1)}{2}-\frac{N_0 (N_0+1)}{2}$ and $H^{d_0}\left(\cF_{M,B_0}\right)=(2 M)^{d_0}$.

Define
\begin{equation*}
    \begin{gathered}
        S_1^{N_0,N} := \left\{(i,j)\mid 1\leq i\leq j\leq N_0\right\},\\
        S_2^{N_0,N} := \left\{(i,j)\mid N_0 < i \leq j \leq N\right\}.
    \end{gathered}
\end{equation*}
Let $\cI_{N_0,N}$ be the set of all functions from $S_2$ to $S_1\cup\{0\}$, i.e.,
\begin{equation*}
    \cI_{N_0,N}:=\left\{I:S_2\mapsto S_1\cup\{0\}\right\}.
\end{equation*}

Let $\bar{\cF}_{M,B_0}^{N_0,N}\subseteq\cF_{M,B_0}$ be the subset of non-increasing LRC, i.e., 
\begin{equation*}
    \bar{\cF}_{M,B_0}^{N_0,N}:=\left\{A\in\cF_{M,B_0}\mid\op{LRC}(A;\cX,N_0)=\op{LRC}(A;\cX,N)\right\}.
\end{equation*}

Then we have $\bar{\cF}_{M,B_0}^{N_0,N} = \bigcup_{I\in\cI_{N_0,N}}\bar{\cF}_{M,B_0}^I$, where
\begin{equation*}
    \bar{\cF}_{M,B_0}^I:=\left\{A\in\cF_{M,B_0}\,\middle|\,\text{for all }(i,j)\in S_2,\
    \begin{cases}
        A_{i,j} = A_{I(i,j)} & \text{if } I(i,j) \neq 0, \\
        A_{i,j} = 0          & \text{otherwise}
    \end{cases}
    \right\}.
\end{equation*}

Since $\dim_{H}\left(\bar{\cF}_{M,B_0}^I\right)=0$ for all $I\in\cI_{N_0,N}$ and $\cI_{N_0,N}$ is finite, by the countable stability of Hausdorff dimension \cite{falconer2013fractal}, we have
\begin{equation*}
    \dim_H\left(\bar{\cF}_{M,B_0}^{N_0,N}\right) = \max_{I\in\cI_{N_0,N}}\left\{\bar{\cF}_{M,B_0}^I\right\} = 0 <d_0.
\end{equation*}

As $\tilde{\cF}_{M,B_0}^{N_0,N}=\cF_{M,B_0}\setminus\bar{\cF}_{M,B_0}^{N_0,N}$, by Lemma~\ref{lm:hausdoff-dim}, we have
\begin{equation*}
    \dim_H\left(\tilde{\cF}_{M,B_0}^{N_0,N}\right) = d_0.
\end{equation*}

For a fixed learning algorithm, the learned parameter is determined by the chosen PRF $\phi$ and the sub-matrix $A_{[N_0],[N_0]}$ of the target. Denote the parameter learned with the PRF $\phi$ and the sub-matrix $B_0\in\cU_{N_0}$ by $A_{\phi,B_0}$. Furthermore, we define
\begin{equation*}
    \cA_{B_0} := \left\{A_{\phi,B_0}\mid \phi\in\Phi_{N,S}\right\},
\end{equation*}
where $\Phi_{N,S}:=\left\{\phi\mid\phi:[N]\times[N]\mapsto[S]\right\}$.
Since $\Phi_{N,S}$ is finite, we have $\dim_H\left(\cA_{B_0}\right)=0 < d_0$.

As $\cF_{M,B_0}^{N_0,N}\subseteq\cA_{B_0}$, we have $\dim_{H}\left(\cF_{M,B_0}^{N_0,N}\right)\leq\dim_{H}\left(A\right)$ and thus $\dim_{H}\left(\cF_{M,B_0}^{N_0,N}\right)=0<d_0=\dim_{H}\left(\tilde{\cF}_{M,B_0}^{N_0,N}\right)$. According to Lemma~\ref{lm:hausdoff-dim}, we have     
\begin{equation*}
    \dim_H\left(\tilde{\cF}_{M,B_0}^{N_0,N}\setminus\cF_{M,B_0}^{N_0,N}\right) = 
    \dim_H\left(\tilde{\cF}_{M,B_0}^{N_0,N}\right).
\end{equation*}
Since $d_N=d_0$, it also holds that 
\begin{equation*}
    H^{d_N}\left(\tilde{\cF}_{M,B_0}^{N_0,N}\setminus\cF_{M,B_0}^{N_0,N}\right) = 
    H^{d_N}\left(\tilde{\cF}_{M,B_0}^{N_0,N}\right).
\end{equation*}

\subsection{Proof for Theorem~\ref{thm:pola-capability}}

As in the proof for Theorem~\ref{thm:pola-limitation}, we define
\begin{equation*}
    \begin{gathered}
        S_1^{N_0,N} := \left\{(i,j)\mid 1\leq i\leq j\leq N_0\right\},\\
        S_2^{N_0,N} := \left\{(i,j)\mid N_0 < i \leq j \leq N\right\}.
    \end{gathered}
\end{equation*}

Let $S=\op{LRC}\left(f^*, \cX_{N_0}\right)+1$.
Since $\op{LRC}\left(f^*, \cX_{N_0}\right)=\op{LRC}\left(f^*, \cX_{N}\right)$, 
there exist $U_1,\cdots,U_{S-1}\in\{0,1\}^{N\times N}$ such that 
$\langle U_s, U_s'\rangle=0$ for $s\neq s'$,
$\left\langle \left(U_s\right)_{[N_0],[N_0]}, \mathds{1}_{N_0\times N_0}\right\rangle > 0$, and
$A^*=\sum_{s=1}^{S-1}a_s U_s, a_s\neq 0$ for all $s=1,\dots,S-1$,
then for each $(i,j)\in S_2^{N_0, N}$, either $A^*_{i,j}=A^*_{i',j'}$ for some $(i',j')\in S_1$ or $A^*_{i,j}=0$. Here, $\mathds{1}_{N_0\times N_0}$ denotes the $N_0\times N_0$ matrix whose entries are all ones.

Define
\begin{equation*}
    \phi(i,j)=\begin{cases}
        s, & \text{if } \left(U_s\right)_{i,j}=1,\\
        S, & \text{otherwise}.
    \end{cases}
\end{equation*}
The PE induced by the PRF $\phi$ is
\begin{equation*}
    A^{\phi} = \sum_{s=1}^S U_s^{\phi} q_s,
\end{equation*}
where $q_1,\dots,q_S$ are learnable parameters and
\begin{equation*}
    \left(U_s^{\phi}\right)_{i,j}=
    \begin{cases}
        1, & \text{if } \phi(i,j)=s,\\
        0, & \text{otherwise},
    \end{cases}
\end{equation*}
for all $s=1,\dots,S$.

We show the POLA model with the above PE initialized at 0 and trained with gradient descent achieves $(N_0,N)$-LG. Notice that $f_{\POLA}\left(x,n;A^{\phi}\right)=\sum_{s=1}^S\left\langle x e_n^\intercal, U_s\right\rangle q_s:=f(q)$ is linear w.r.t. to the learnable parameters $ q = \left[q_1,\dots,q_S\right]^\intercal$. Then the learned interpolator $f(\cdot,\cdot;\hat{q})$ is that minimizes the $\ell_2$-norm of $q$ \cite{bartlett2021deep}, i.e.,
\begin{equation*}
    \hat{a} = \argmin \|q\|_2, \text{ s.t. } f(x,n;q)=f_{\POLA}\left(x,n;A^*\right) \text{ for all } x\in\cX, n\in [N_0].
\end{equation*}

As $f\left(\cdot,\cdot;\hat{q}\right)$ is an interpolator for all $x\in\cX, n\in [N_0]$, we have
\begin{equation*}
    \left(\sum_{s=1}^S U_s \hat{q}_s\right)_{[N_0],[N_0]} 
    = A^*_{[N_0],[N_0]} 
    = \left(\sum_{s=1}^S U_s a_s\right)_{[N_0],[N_0]}
\end{equation*}

For $s=1,\dots,S-1$, we have
\begin{equation*}
    \left\langle \left(\sum_{s'=1}^S U_{s'} \hat{q}_{s'}\right)_{[N_0],[N_0]}, \left(U_s\right)_{[N_0],[N_0]} \right\rangle
    =
    \left\langle \left(\sum_{s'=1}^S U_{s'} a_{s'}\right)_{[N_0],[N_0]}, \left(U_s\right)_{[N_0],[N_0]}\right\rangle,
\end{equation*}
which implies
\begin{equation*}
    \left\langle \left(U_s\right)_{[N_0],[N_0]}, \left(U_s\right)_{[N_0],[N_0]} \right\rangle \hat{q}_s
    =
    \left\langle \left(U_s\right)_{[N_0],[N_0]}, \left(U_s\right)_{[N_0],[N_0]} \right\rangle a_s.
\end{equation*}

As $\left\langle \left(U_s\right)_{[N_0],[N_0]}, \left(U_s\right)_{[N_0],[N_0]} \right\rangle > 0$, we have $\hat{q}_s=a_s$ for all $s=1,\dots,S-1$. Hence,
\begin{equation*}
    \left\{q\mid f(x,n;q)=f_{\POLA}\left(x,n;A^*\right) \text{ for all } x\in\cX, n\in [N_0]\right\} \subseteq \{q\mid q_s = a_s \text{ for all } s=1,\dots,S-1\}:=Q.
\end{equation*}

Notice that when $\hat{q}_S=0$, the function $f(\cdot,\cdot;\hat{q})$ interpolates the training data, and $\hat{q}=\argmin_{q\in Q}\|q\|_2$. Therefore, the learned model is $f(\cdot,\cdot;\hat{q})$ with $\hat{q}=\left[a_1,\dots, a_{S-1}, 0\right]$. Since $f(x,n;\hat{q})=f_{\POLA}(x,n;A^*)$ for all $x\in\cX, n\in[N_0]$, the model achieves $(N_0,N)$-LG.

\subsection{Proof for Theorem~\ref{thm:limitation}}

Define
\begin{equation*}
    \cF_{N_0}:=\left\{f_0\,\middle|\, f_0:\{0,1\}^{[N_0]}\mapsto\{0,1\}^K\right\},
    \Phi_{N,S}:=\left\{\phi:[N]\times[N]\mapsto[S]\right\}.
\end{equation*}

For a fixed learning algorithm, the learned model is determined by the target function restricted on $\{0,1\}^{[N_0]}$ and the PE. In other words, each pair $\left(u, \phi\right)\in\cF_{N_0}\times\Phi_{N,S}$ corresponds to a model (denoted by $f_{u, \phi}$) learned by the algorithm. Let $\cF'_{N_0,\Phi_{N,S}}$ be the set of all these models, i.e.,
\begin{equation*}
    \cF'_{N_0,\Phi_{N,S}} := \left\{f_{u,\phi}\mid u\in\cF_{N_0}, \phi\in\Phi_{N,S}\right\}.
\end{equation*}

We have
\begin{equation*}
    |\cF_{N_0,N}| \leq \left|\cF'_{N_0,\Phi_{N,S}}\right| \leq |\cF_{N_0}| |\Phi_{N,S}| = 2^{K\sum_{n=1}^{N_0} 2^{n}} \times S^{N^2}=2^{K \left(2^{N_0 + 1}-1\right)}\times S^{N^2}.
\end{equation*}

Let $\bar{\cF}_{N_0,N}\subseteq\cF_N$ be the subset of functions with non-increasing SRC, i.e.,
\begin{equation*}
    \bar{\cF}_{N_0,N}:=\left\{f\in\cF_N\mid\op{SRC}(f,N_0)=\op{SRC}(f,N)\right\}.
\end{equation*}

We have
\begin{equation*}
    \left|\bar{\cF}_{N_0,N}\right|
    \leq \left|\cF_{N_0}\right| \sum_{n_0=N_0}^{N-1}\sum_{n=1}^{N_0} \binom{n_0}{n}
    = 2^{K \left(2^{N_0 + 1}-1\right)}\times \sum_{n_0=N_0}^{N-1}\sum_{n=1}^{N_0} \binom{n_0}{n}.
\end{equation*}

Since 
\begin{equation*}
    \frac{\left|\tilde{\cF}_{N_0,N}\setminus\cF_{N_0,N}\right|}{\left|\tilde{\cF}_{N_0,N}\right|}
    \ge
    \frac{\left|\cF_N\right|-\left|\bar{\cF}_{N_0,N}\right|-\left|\cF_{N_0,N}\right|}{\left|\cF_N\right|-\left|\bar{\cF}_{N_0,N}\right|},
\end{equation*}
and $\left|\cF_{N_0,N}\right|$, $\left|\bar{\cF}_{N_0,N}\right|$ are monotonously increasing w.r.t. $N_0$, it suffices to consider $N_0=N-1$. Then we have
\begin{equation*}
    \begin{gathered}
        \left|\cF_{N_0,N}\right|\leq 2^{K\left(2^N-1\right)}\times S^{N^2},\\
        \left|\bar{\cF}_{N_0,N}\right|\leq 2^{K \left(2^N-1\right)}\times \sum_{n_0=N_0}^{N-1}\sum_{n=1}^{N_0} \binom{n_0}{n}=2^{K \left(2^N-1\right)}\times \left(2^N - 1\right).
    \end{gathered}
\end{equation*}
Noting that 
\begin{equation*}
    \left|\cF_N\right| = 2^{K\sum_{n=1}^{N}=2^n} = 2^{K \left(2^{N+1}-1\right)},
\end{equation*}
we have
\begin{equation*}
    \begin{aligned}
        1 \ge \frac{\left|\tilde{\cF}_{N_0,N}\setminus\cF_{N_0,N}\right|}{\left|\tilde{\cF}_{N_0,N}\right|} 
          &\ge \frac{2^{K \left(2^{N+1}-1\right)} - 2^{K \left(2^N-1\right)}\times \left(2^N - 1\right) - 2^{K \left(2^N-1\right)}\times S^{N^2}}{2^{K \left(2^{N+1}-1\right)}-2^{K \left(2^N-1\right)}\times \left(2^N - 1\right)}\\
          &=\frac{2^{K 2^N}-\left(2^N-1\right)-S^{N^2}}{2^{K 2^N}-\left(2^N-1\right)}\\
          &=1-\frac{S^{N^2}}{2^{K 2^N}-\left(2^N-1\right)}.
    \end{aligned}
\end{equation*}
As $N\to\infty$, we obtain 
\begin{equation*}
    \lim_{N\to\infty}\frac{\left|\tilde{\cF}_{N_0,N}\setminus\cF_{N_0,N}\right|}{\left|\tilde{\cF}_{N_0,N}\right|} = 1.
\end{equation*}

\subsection{Proof for Theorem~\ref{thm:scale-hint}}

Consider a circuit representation $\cC=\{C_n\}$ of non-increasing SRC and denote the corresponding mapping by $f$. Then we have $\op{SRC}(f,N_0)=\op{SRC}(f,N)$.

By the definition of SRC, there exists a set of operators $\cG=\{g_1,\dots, g_K\}$, where $g_k:\Sigma^{n_k}\mapsto\Sigma, n_k \leq N_0$ for all $k=1,\dots,K$.

Define the PRF-SH
\begin{equation*}
    \phi(i,j,n):=\begin{cases}
        \sum_{k=1}^{k'-1} n_k + I^{-1}_i(j), &\text{ if } \left(i, g_{k'}, I_i\right)\in C_n,\\
        \sum_{k=1}^{K} n_k, & \text{otherwise}.
    \end{cases}
\end{equation*}

We will show that the PRF-SH $\phi(i,j,n)$ characterizes $\cC$. The consistency follows directly from the definition of $\phi$. It remains to check the distinctness. 

For any distinct pairs $(i,j,m)\neq (i',j',n)$, suppose that $(i,g^{(i)},I_i)\in C_m$ and $j\in I_i$. Then for $(i',g^{(i')}, I_{i'})\in C_n$, we have:
\begin{itemize}
    \item When $j'\in I_{i'}$, we have:
    \begin{itemize}
        \item If $g^{(i)}\neq g^{(i')}$ (without loss of generality, we assume $g^{(i)}=g_{k_1}$, $g^{(i')}=g_{k_2}$, and $k_1 < k_2$), then
        \begin{equation*}
            \phi(i,j,m) = \sum_{k=1}^{k_1-1} n_k + I^{-1}_i(j) 
            \leq \sum_{k=1}^{k_1} n_k
            < \sum_{k=1}^{k_1} n_k + 1
            \leq \sum_{k=1}^{k_2-1} n_k + I_{i'}^{-1}(j')
            =\phi(i',j',n);
        \end{equation*}
        \item If $g^{(i)}= g^{(i')}=g_{k'}$ but $I_{i}^{-1}(j)\neq I_{i'}^{-1}(j')$, then 
        \begin{equation*}
            \phi(i,j,m) = \sum_{k=1}^{k'-1} + I_{i}^{-1}(j) \neq \sum_{k=1}^{k'-1} + I_{i'}^{-1}(j') = \phi(i',j',n).
        \end{equation*}
    \end{itemize}
    \item When $j'\not\in I_{i'}$, we have
    \begin{equation*}
        \phi(i,j,m)<+\infty=\phi(i',j',n).
    \end{equation*}
\end{itemize}
Therefore, the PRF-SH $\phi$ satisfies the distinctness condition.
\section{Experimental Details}\label{appdx:experiments}

\textbf{Data.} When the scale hint is not applied, we align all the instances (except for Select) to the maximum scale by filling with ``0''. This is to guarantee the existence of characterizing PRFs for the task. For example, for the copy task where the training instances are of scales 1--5 and the testing instances are of scales 6--10, we align all instances to scale 10. An aligned training instance is like
\begin{equation*}
    \mathtt{x_1\; x_2\; x_3\; x_4\; x_5\; 0\; 0\; 0\; 0\; 0\;=\;x_1\; x_2\; x_3\; x_4\; x_5\; 0\; 0\; 0\; 0\; 0}.
\end{equation*}
When the scale hint is applied, we do not align the instances because we can always find a characterizing PRF-SH according to Theorem~\ref{thm:scale-hint}. An unaligned copy instance of scale $n$ is like
\begin{equation*}
    \mathtt{x_1\; \dots\; x_n\;=\;x_1\; \dots\; x_n\;}.
\end{equation*}
We summarize the data formats of the tasks in our experiments in Table~\ref{tab:data-format}.

\begin{table}[ht]
    \centering
    \caption{Data formats of the tasks in our experiments.}
    \small
    \begin{tabular}{lcc}
    
    \toprule
    
    Task & Unaligned format & Aligned format \\
    
    \midrule
    
    Copy & $\mathtt{x_1\;\dots\;x_n\;=\;x_1\;\dots\;x_n}$ & $\mathtt{x_1\;\dots\; x_n\;0\;\dots\;0\;=\;x_1\;\dots\;x_n\;0\;\dots\;0}$ \\
    Reverse & $\mathtt{x_1\;\dots\;x_n\;=\;x_n\;\dots\;x_1}$ & $\mathtt{x_1\;\dots\;x_n\;0\;\dots\;0\;=\;0\;\dots\;0\;x_n\;\dots\;x_1}$ \\
    Shift & $\mathtt{x_1\;x_2\;\dots\;x_n\;=\;x_2\;\dots\;x_n\;x_1}$ & $\mathtt{x_1\;x_2\;\dots\;x_n\;0\;\dots\;0\;=\;x_2\;\dots\;x_n\;0\;\dots\;0\;x_1}$\\
    Parity (with CoT) & $\mathtt{x_1\;\dots\;x_n\;=\;y_1\;\dots\;y_n}$ & $\mathtt{x_1\;\dots\;x_n\;0\;\dots\;0\;=\;y_1\;\dots\;y_n\;y_n\;\dots\;y_n}$ \\
    Addition & $\mathtt{x_1\;\dots\;x_n\;+\;y_1\;\dots\;y_n\;=\;z_1\;\dots\;z_n}$ & 
    $\mathtt{x_1\;\dots\;x_n\;0\;\dots\;0+\;y_1\;\dots\;y_n\;0\;\dots\;0=\;z_1\;\dots\;z_n\;0\;\dots\;0}$\\
    Multiplication (1 * N) & $\mathtt{y_1\;*\;x_1\;\dots\;x_n\;=\;z_1\;\dots\;z_n\;z_{n+1}}$ & 
    $\mathtt{y_1\;*\;x_1\;\dots\;x_n\;0\;\dots\;0\;=\;z_1\;\dots\;z_n\;z_{n+1}\;0\;\dots\;0}$\\
    Division (N / 1) & $\mathtt{y_1\;\backslash\;x_n\;\dots\;x_1\;=\;z_n\;\dots\;z_1}$ & $\mathtt{y_1\;\backslash\;0\;\dots\;0\;x_n\;\dots\;x_1\;=0\;\dots\;0\;z_n\;\dots\;z_1}$\\
    Select & $\mathtt{x_1\;\dots\;x_n\;=y}$ & \textemdash \\
    \bottomrule
    \end{tabular}
    
    \label{tab:data-format}
\end{table}

To reduce the data requirements, all Addition instances are in base 3.

\textbf{Models.} We train GPT-2 models with various PEs. We summarize the PRFs of the IPEs and the PRF-SHs of the IPE-SHs used in the experiments in Table~\ref{tab:prf-prfsh}.

\begin{table}[ht]
    \centering
    \caption{PRFs for the IPEs and PRF-SHs for the IPE-SHs in our experiments.}
    \small
    \begin{tabular}{lll}
    
    \toprule
    
    Task & PRF & PRF-SH \\

    \midrule
    Copy &  \scriptsize $\phi(i,j)=\left\{\begin{array}{ll}
            1, & \text{ if } i - j = 20,\\
            0, & \text{ otherwise}.
            \end{array}\right.$  & \textemdash \\

    Shift & \scriptsize $\phi(i,j)=\left\{\begin{array}{ll}
            1, & \text{ if } i = 2N-1 \text{ and } j = 0,\\
            2, & \text{ if } i < 2N-1 \text{ and } i-j = N-1,\\
            0, & \text{ otherwise}.
            \end{array}\right.$  & \textemdash \\
    Parity (with CoT) & \scriptsize $\phi(i,j)=\left\{\begin{array}{ll}
            1, & \text{ if } i - j = 0,\\
            2, & \text{ if } i - j = N,\\
            0, & \text{ otherwise}.
            \end{array}\right.$  & \textemdash \\
    Addition & \scriptsize $\phi(i,j)=\left\{\begin{array}{ll}
            1, & \text{ if } i - j = 0,\\
            2, & \text{ if } i - j = N,\\
            3, & \text{ if } i - j = N+1,\\
            4, & \text{ if } i - j = 2N+1,\\
            5, & \text{ if } i - j = 2N+2,\\
            0, & \text{ otherwise}.
            \end{array}\right.$  & \scriptsize $\phi(i,j,n)=\left\{\begin{array}{ll}
            1, & \text{ if } i - j = 0,\\
            2, & \text{ if } i - j = n,\\
            3, & \text{ if } i - j = n+1,\\
            4, & \text{ if } i - j = 2n+1,\\
            5, & \text{ if } i - j = 2n+2,\\
            0, & \text{ otherwise}.
            \end{array}\right.$ \\
    Multiplication (1 * N) & \scriptsize $\phi(i,j)=\left\{\begin{array}{ll}
            1, & \text{ if } j = 0,\\
            2, & \text{ if } i - j = 0,\\
            3, & \text{ if } i - j = N,\\
            4, & \text{ if } i - j = N+1,\\
            0, & \text{ otherwise}.
            \end{array}\right.$  & \scriptsize $\phi(i,j,n)=\left\{\begin{array}{ll}
            1, & \text{ if } j = 0,\\
            2, & \text{ if } i - j = 0,\\
            3, & \text{ if } i - j = n,\\
            4, & \text{ if } i - j = n+1,\\
            0, & \text{ otherwise}.
            \end{array}\right.$ \\
    Division (N / 1) & \scriptsize $\phi(i,j)=\left\{\begin{array}{ll}
            1, & \text{ if } j = 0,\\
            2, & \text{ if } i - j = 0,\\
            3, & \text{ if } i - j = N,\\
            4, & \text{ if } i - j = N+1,\\
            0, & \text{ otherwise}.
            \end{array}\right.$  & \scriptsize $\phi(i,j,n)=\left\{\begin{array}{ll}
            1, & \text{ if } j = 0,\\
            2, & \text{ if } i - j = 0,\\
            3, & \text{ if } i - j = n,\\
            4, & \text{ if } i - j = n+1,\\
            0, & \text{ otherwise}.
            \end{array}\right.$ \\
    \bottomrule
    \end{tabular}
    \label{tab:prf-prfsh}
\end{table}

\textbf{Training.} All the models are trained with AdamW and the same experiment groups share the hyperparameters.

Settings of the dataset sizes, the model hyperparameters, and the training recipes are listed in Tables~\ref{tab:es:ipe-ape-rpe}--\ref{tab:es:lbpe-lbpesh}.

\begin{table}[ht]
    \centering
    \begin{minipage}{0.49\textwidth}
    \centering
    \caption{Settings for the experiments in Fig.~\ref{fig:ipe-ape-rpe}.}
    \small
    \begin{tabular}{lll}
    
    \toprule
    
    \multirow{2}{*}{Data} 
        & Number of training examples & 10,000 \\
        & Number of testing examples & 1,000 \\
    \midrule
    
    \multirow{5}{*}{Model} 
        & Base model & GPT-2 \\
        & Number of layers & 6 \\
        & Number of attention heads & 1 \\
        & Hidden dimension & 768 \\
        & Number of PRF values & 128 \\
    \midrule
    
    \multirow{7}{*}{Training Recipe} 
        & Effective batch size & 1,024 \\
        & Number of epochs & 300 \\
        & Optimizer & AdamW \\
        & Learning rate & $5 \times 10^{-4}$ \\
        & Weight decay & 1.0 \\
        & Warmup ratio & 0.05 \\
        & Learning rate scheduler & cosine \\
    
    \bottomrule
    \end{tabular}
    \label{tab:es:ipe-ape-rpe}
    \end{minipage}
    \hfill
    \begin{minipage}{0.49\textwidth}
    \centering
    \caption{Settings for the experiments in Fig.~\ref{fig:scale-hint}.}
    \small
    \begin{tabular}{lll}
    
    \toprule
    
    \multirow{2}{*}{Data} 
        & Number of training examples & 10,000 \\
        & Number of testing examples & 1,000 \\
    \midrule
    
    \multirow{5}{*}{Model} 
        & Base model & GPT-2 \\
        & Number of layers & 6 \\
        & Number of attention heads & 1 \\
        & Hidden dimension & 768 \\
        & Number of PRF values & 128 \\
    \midrule
    
    \multirow{7}{*}{Training Recipe} 
        & Effective batch size & 1,024 \\
        & Number of epochs & 300 \\
        & Optimizer & AdamW \\
        & Learning rate & $5 \times 10^{-4}$ \\
        & Weight decay & 1.0 \\
        & Warmup ratio & 0.05 \\
        & Learning rate scheduler & cosine \\
    
    \bottomrule
    \end{tabular}
    \label{tab:es:scale-hint}
    \end{minipage}
    \vskip 1em 
    \begin{minipage}{0.49\textwidth}
    \centering
    \caption{Settings for the experiments in Fig.~\ref{fig:lbpe}.}
    \small
    \begin{tabular}{lll}
    
    \toprule
    
    \multirow{2}{*}{Data} 
        & Number of training examples & 1,000 \\
        & Number of testing examples & 1,000 \\
    \midrule
    
    \multirow{5}{*}{Model} 
        & Base model & GPT-2 \\
        & Number of layers & 1 \\
        & Number of attention heads & 1 \\
        & Hidden dimension & 768 \\
        & Number of PRF values & 64 \\
    \midrule
    
    \multirow{7}{*}{Training Recipe} 
        & Effective batch size & 1,024 \\
        & Number of epochs & 2,000 \\
        & Optimizer & AdamW \\
        & Learning rate & $5 \times 10^{-4}$ \\
        & Weight decay & 1.0 \\
        & Warmup ratio & 0.05 \\
        & Learning rate scheduler & cosine \\
    
    \bottomrule
    \end{tabular}
    
    \label{tab:es:lbpe}
    \end{minipage}
    \hfill
    \begin{minipage}{0.49\textwidth}
    \centering
    \caption{Settings for the experiments in Fig.~\ref{fig:lbpe-lbpesh}.}
    \small
    \begin{tabular}{lll}
    
    \toprule
    
    \multirow{2}{*}{Data} 
        & Number of training examples & 10,000 \\
        & Number of testing examples & 1,000 \\
    \midrule
    
    \multirow{5}{*}{Model} 
        & Base model & GPT-2 \\
        & Number of layers & 6 \\
        & Number of attention heads & 1 \\
        & Hidden dimension & 768 \\
        & Number of PRF values & 128 \\
    \midrule
    
    \multirow{7}{*}{Training Recipe} 
        & Effective batch size & 1,024 \\
        & Number of epochs & 10,000 \\
        & Optimizer & AdamW \\
        & Learning rate & $5 \times 10^{-4}$ \\
        & Weight decay & 1.0 \\
        & Warmup ratio & 0.05 \\
        & Learning rate scheduler & cosine \\
    
    \bottomrule
    \end{tabular}
    
    \label{tab:es:lbpe-lbpesh}
    \end{minipage}
\end{table}

\end{document}